\pdfoutput=1

\documentclass[11pt]{article}

\usepackage[]{EMNLP2023}

\usepackage{times}
\usepackage{latexsym}

\usepackage[T1]{fontenc}

\usepackage[utf8]{inputenc}

\usepackage{microtype}

\usepackage{inconsolata}
\usepackage{marvosym} 

\usepackage{graphicx}
\usepackage{graphicx}
\usepackage{amsmath,amssymb,amsthm,amsfonts}
\usepackage{bbm}
\usepackage{acronym}
\usepackage{enumitem}
\usepackage{balance}
\usepackage{xspace}
\usepackage[skip=3pt]{subcaption}
\usepackage[skip=3pt]{caption}
\usepackage[switch]{lineno}
\usepackage{url}
\usepackage{multirow}
\usepackage{booktabs}
\usepackage{mathrsfs}
\usepackage{arydshln}
\usepackage{array}
\usepackage{colortbl}
\usepackage{pifont}

\definecolor{myblue}{RGB}{0,105,252}
\definecolor{myred}{RGB}{247,96,102}
\definecolor{mygray}{gray}{0.4}

\makeatletter
\DeclareRobustCommand\onedot{\futurelet\@let@token\@onedot}
\def\@onedot{\ifx\@let@token.\else.\null\fi\xspace}
\def\eg{\emph{e.g}\onedot}

\def\ie{\emph{i.e}\onedot}

\def\etc{\emph{etc}\onedot} 

\def\wrt{w.r.t\onedot}

\makeatother

\acrodef{nlp}[NLP]{natural language processing}
\acrodef{plm}[PLM]{pretrained language model}
\acrodef{sota}[SOTA]{state-of-the-art}
\acrodef{bs}[BS]{Beam Search}
\acrodef{mhs}[MHS]{Metropolis-Hastings Sampling}
\acrodef{hs}[HS]{Hybrid Search}
\acrodef{uas}[UAS]{unlabeled attachment score}
\acrodef{dda}[DDA]{Directed Dependency Accuracy}
\acrodef{sota}[SOTA]{state-of-the-art}
\acrodef{pos}[POS]{part-of-speech}
\acrodef{asr}[ASR]{attacking success rate}
\acrodef{ppl}[PPL]{Perplexity score}
\acrodef{cqr}[CQR]{Conversational Question Reformulation}
\acrodef{cqa}[CQA]{Conversational Question Answering}
\acrodef{mcqr}[MTCQR]{Multi-Topic Conversational Question Reformulation}
\acrodef{amt}[AMT]{Amazon Mechanical Turk}
\acrodef{mtcl}[MTCL]{Multi-Topic Contrastive Learning}

\newcommand{\model}{{\fontfamily{lmtt}\selectfont CollabKG}\xspace}
\title{CollabKG: A Learnable Human-Machine-Cooperative Information Extraction Toolkit for (Event) Knowledge Graph Construction}

\author{Xiang Wei$^{1}$, 
Yufeng Chen$^{1}$, 
Ning Cheng$^{1}$, 
Xingyu Cui$^{1}$,  
Jinan Xu$^{1}$, 
and Wenjuan Han$^{1\,\textrm{\Letter}}$ \\
\textsuperscript{1} Beijing Jiaotong University, Beijing, China \\
}

\begin{document}
\maketitle
\begin{abstract}
In order to construct or extend entity-centric and event-centric knowledge graphs (KG and EKG), the information extraction (IE) annotation toolkit is essential. However, existing IE toolkits have several non-trivial problems, such as not supporting multi-tasks, not supporting automatic updates. In this work, we present \model, a learnable human-machine-cooperative IE toolkit for KG and EKG construction. Specifically, for the multi-task issue, \model unifies different IE subtasks, including named entity recognition (NER), entity-relation triple extraction (RE), and event extraction (EE), and supports both KG and EKG. Then, combining advanced prompting-based IE technology, the human-machine-cooperation mechanism with LLMs as the assistant machine is presented which can provide a lower cost as well as a higher performance. Lastly, owing to the two-way interaction between the human and machine, \model with learning ability allows self-renewal. Besides, \model has several appealing features (\eg, customization, training-free, propagation, \etc) that make the system powerful, easy-to-use, and high-productivity. We holistically compare our toolkit with other existing tools on these features. Human evaluation quantitatively illustrates that \model significantly improves annotation quality, efficiency, and stability simultaneously.~\footnote{\scriptsize Github: \url{https://github.com/cocacola-lab/CollabKG} \\Video: \url{https://www.youtube.com/channel/UCsadiRvhW9dsmn4KtRDCaFg} \\Demo: \url{https://github.com/cocacola-lab/CollabKG}}
\end{abstract}

\section{Introduction}\label{sec:introduction}
Entity-centric and event-centric knowledge graphs (KG and EKG) are structured semantic knowledge bases for describing concepts and their relationships in the physical world~\citep{zou2020survey, guan2022event}. These KGs, along with EKGs (collectively referred to as (E)KGs), are playing an increasingly important role in many downstream tasks and applications, including search engine~\citep{zhao2021brain,yang2020temporal}, recommendation systems~\citep{wang2019kgat}, question-answering~\citep{bao2016constraint, souza2020event}, commonsense reasoning~\citep{lin2019kagnet} and text generation~\citep{yang2020temporal}. 
With the dynamic changes in the content of the Internet, existing (E)KGs still need to be completed in the general domains, and even need to be constructed from scratch in emerging and specialized domains~\citep{kejriwal2022knowledge,chen2020knowledge}.
In this regard, the information extraction (\ie, IE) technique is an effective way to construct or complement (E)KGs~\citep{luan2018multi,li2020real}.

\begin{figure*}[htp]
\centering
\includegraphics[width=1\linewidth]{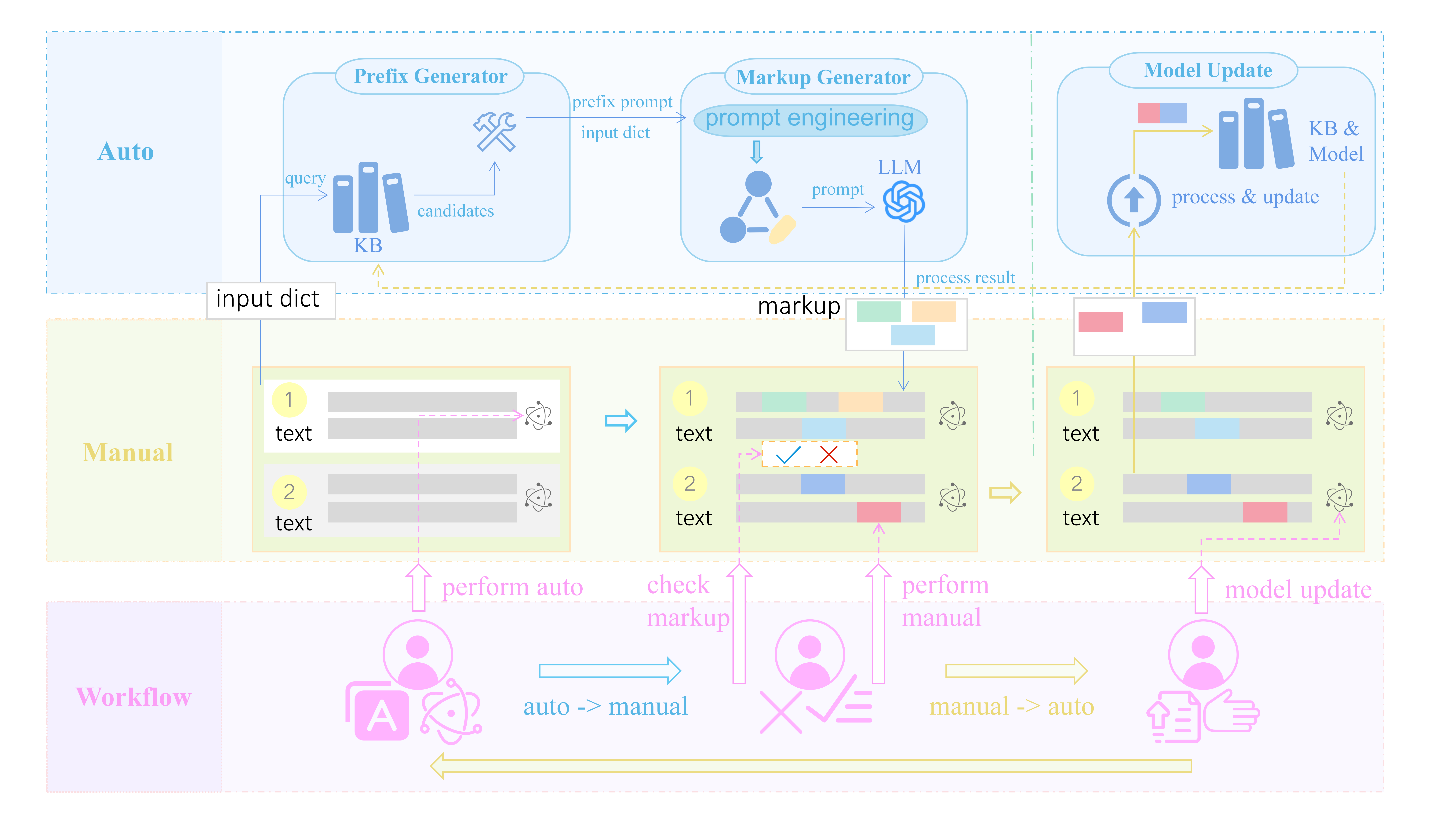}  
\caption{Illustration of the human-machine-cooperative workflow for \model, as well as how automatic and manual labeling modules collaborate.}
\label{fig:interaction}
\vspace{-5mm}
\end{figure*}

There are tons of existing open-source tools for IE labeling, both automatically and manually. However, these tools still have some non-trivial issues that hinder the applicability and effectiveness of real-world applications. First, there are various IE tasks, such as named entity recognition (NER), entity-relation triple extraction (RE), and event extraction (EE). However, most tools only support one or two of these tasks~\citep{nghiem-ananiadou-2018-aplenty,10.1093/nar/gkaa333,stewart-etal-2019-redcoat,li-etal-2021-fitannotator}. To my knowledge, very few open-source manual annotation toolkits function EE annotation. Hence, KG and EKG often use individual tools instead of a unified one while parts of KG and EKG share similar architecture. Secondly, most tools only support either automatic or manual labeling~\citep{stenetorp-etal-2012-brat,zhang2022deepke,bikaun2022quickgraph,jin2021cogie,9378107}. However, a human-machine cooperative system has proven to outperform both standalone agents and humans working alone~\citep{bien2018deep}. Lastly, even if some tools support both automatic and manual labeling, the machine itself cannot learn from human annotations as feedback~\citep{klie-etal-2018-inception,inproceedings}.
Besides the above three major issues, minor issues exist like requiring large amounts of data for training, which makes labeling time-consuming and not suitable for low-resource scenarios~\citep{jin2021cogie}.
Therefore, it is crucial to build an IE toolkit that is multi-tasking, human-machine-cooperative, training-free, \etc.

Based on the above clues, we propose \model (See Fig.~\ref{fig:interaction}), a learnable human-machine-cooperative IE annotation toolkit for KG and EKG construction. The main contributions are described as follows:
\begin{itemize}
    \item[1] \model is an open-source IE annotation toolkit that unifies NER, RE, and EE tasks, integrates KG and EKG, and supports both English and Chinese languages. 
    \item[2] \model combines automatic and manual labeling to build a learnable human-machine cooperative system. In particular, humans benefit from machines and meanwhile, manual labeling provides a reference for machines' self-renewal in an online way. Additionally, \model is designed with many other appealing features that are conducive to productivity, powerful, and user-friendly.
    \item[3] Extensive human evaluation suggest that \model can significantly improve the effectiveness and efficiency of manual annotation, as well as reduce variance (Sec.~\ref{sec:experiments}).
\end{itemize}

\section{Core Functions}\label{sec:core functions}

\begin{table*}[!htb]
    \centering
    \resizebox{0.98\textwidth}{!}{
    \begin{tabular}{cccccccccc}
    \toprule
          & \textbf{NER} & \textbf{RE} & \textbf{EE} & \textbf{Auto} & \textbf{Manual} & \textbf{Learn} & \textbf{TF.} & \textbf{AP.} & \textbf{DC.}\\
    \midrule
       \textbf{\model} & \checkmark & \checkmark & \checkmark & \checkmark & \checkmark & \checkmark & \checkmark & \checkmark & \checkmark\\
        DeepKE~\citep{zhang2022deepke} & \checkmark & \checkmark & - & \checkmark & - & - & - & - & -\\
        CogIE~\citep{jin2021cogie} & \checkmark & \checkmark & \checkmark & \checkmark & - & - & - & - & -\\
        Quickgraph~\citep{bikaun2022quickgraph} & \checkmark & \checkmark & - & - & \checkmark & - & - & \checkmark & \checkmark\\
        BRAT~\citep{stenetorp-etal-2012-brat} & \checkmark & \checkmark & \checkmark & - & \checkmark & - & - & - & -\\
        WebAnno~\citep{yimam-etal-2013-webanno} & \checkmark & - & - & - & \checkmark & - & - & - & -\\
        SLATE~\citep{kummerfeld-2019-slate} & \checkmark & \checkmark & \checkmark & - & \checkmark & - & - & - & -\\
        INCEpTION~\citep{klie-etal-2018-inception} & \checkmark & \checkmark & - & \checkmark & \checkmark& - & - & - & -\\
        TeamTat~\citep{10.1093/nar/gkaa333} & \checkmark & \checkmark & - & - & \checkmark & - & - & \checkmark & -\\
        TextAnnotator~\citep{inproceedings} & \checkmark & \checkmark & - & \checkmark & \checkmark & - & - & - & -\\
        FITAnnotator~\citep{li-etal-2021-fitannotator} & \checkmark & - & - & \checkmark & \checkmark & \checkmark & - & - & -\\
        APLenty~\citep{nghiem-ananiadou-2018-aplenty} & \checkmark & - & - & \checkmark & \checkmark & \checkmark & - & - & -\\
        Redcoat~\citep{stewart-etal-2019-redcoat} & \checkmark & - & - & - & \checkmark & - & - & - & -\\
        SALKG~\citep{9378107} & \checkmark & \checkmark & - & - & \checkmark & - & - & - & -\\
        RESIN~\citep{wen-etal-2021-resin} & \checkmark & \checkmark & \checkmark & \checkmark & - & - & - & - & -\\
        REES~\citep{aone-ramos-santacruz-2000-rees} & - & \checkmark & \checkmark & - & \checkmark& - & - & - & -\\
        FLAIR~\citep{akbik-etal-2019-flair} & \checkmark & - & - & \checkmark & - & - & - & - & -\\
        OpenNRE~\citep{han-etal-2019-opennre} & - & \checkmark & - & \checkmark & - & - & - & - & -\\
        ODIN~\citep{valenzuela-escarcega-etal-2015-domain} & - & - & \checkmark & \checkmark & - & - & \checkmark & - & -\\
    \bottomrule
    \end{tabular}
    }
    \caption{Illustration of core functions for \model and comparison with recent existing open-source IE toolkits. \textbf{Auto}: Automatic labeling. \textbf{Manual.}: Manual labeling. \textbf{Learn}: Learnable. \textbf{TF.}: Train free. \textbf{AP.}: Annotation Propagation. \textbf{DC.}: Document Clustering.}
    \label{tab:comparison}
    \vspace{-5mm}
\end{table*}

We compare \model quantitatively with existing open-source IE annotations tools in Tab. \ref{tab:comparison} and perform the following summary about the functions:

\paragraph{Unification:} 83\% of the reviewed tools support NER, 66\% support RE, and 33\% support EE. Only 22\% of the tools support all three IE tasks. 
Moreover, it supports KG and EKG in both English and Chinese.

\paragraph{Human-machine-cooperation: } 55\% and 66\% of the reviewed tools support automatic labeling and manual labeling, respectively. 22\% of tools support both, but they do not support learnability or unification feature. 

\paragraph{Learnability:}
Human labeling can provide a reference for the machine to help machines to annotate more effectively and efficiently.

\paragraph{Other functions:} Only our toolkit \model along with Quickgraph supports annotation propagation and document clustering\footnote{Annotation propagation: Case-insensitive sub-string matching to perform relation/entity propagation \citep{bikaun2022quickgraph}. Document clustering: Clustering documents to promote annotator consistency.}. But the annotation propagation of Quickgraph cannot be applied to Chinese texts. Additionally, \model also has other appealing functions. For example, \model utilizes a prompt-based automatic annotation algorithm for zero-shot IE through prompting large language models (LLMs), which makes it training-free, suitable for low-resource scenarios, and flexible for customization. 
To sum up, \model is designed with many features that are conducive to productivity, powerful, and user-friendly.


\section{Design and Implementation}\label{sec:design and implementation}

\subsection{Implementation of Unification}\label{sec:unify Task}
\model is designed to unify three IE tasks including NER, RE, and EE, and integrate KG and EKG construction (Appendix \ref{sec:implementation_of_unification_all}). 
To unify these tasks, we first observe their schemes and summarize the transformation rules among them. Then we uniformly integrate these three schemes utilizing these transformation rules and design an annotation format to fit the unified scheme.

\paragraph{Transformation Rule}\label{sec: tasks description}
NER aims to find entities with specific preset entity types from the given text. For example, given the preset type list \texttt{[PER, LOC, ORG, MISC]} and the sentence  ``\textit{James worked for Google in Tokyo, the capital of Japan.}'', these entities should be recognized are: \texttt{PER:James, ORG:Google, LOC:Tokyo, LOC:Japan}.

RE aims to find pairs of entities, predict the relations between them and form triples.
For instance, given the sentence ``\textit{Mr.Johnson retired before the 2005 season and briefly worked as a football analyst for WBZ-TV in Boston.}'', this triple is \texttt{(Person:Mr.Johnson, person-company, Organization:WBZ-TV)}.
The first term in the triple is called \texttt{Subject}, the middle term is \texttt{Relation}, and the last term is called \texttt{Object}.

EE plays an important role in EKG construction. It aims to identify event types, triggers, arguments involved, and the corresponding roles. For instance, given the sentence ``\textit{Yesterday Bob and his wife got married in Beijing.}'', we regard \texttt{Life:Marry} as event type, ``\textit{married}'' as the trigger, ``\textit{Bob and his wife}'' as \texttt{Person}, ``\textit{Yesterday}'' as \texttt{Time} and ``\textit{Beijing}'' as \texttt{Place}.   

\begin{table}[!htb]
    \centering
    \resizebox{0.85\columnwidth}{!}{
    \begin{tabular}{c|c}
    \toprule
         NER &  \begin{tabular}[c]{@{}c@{}}\texttt{E-Type:E} \\ $\longrightarrow$\texttt{(E-Type:E, \_, \_)}\end{tabular}  \\ \hline
        RE & \texttt{(S-Type:S, R, O-Type:O)} \\ \hline
        EE & \begin{tabular}[c]{@{}c@{}}\texttt{$\{$E-Type:T, $R_1$:$A_1$, ..., $R_n$:$A_n$ $\}$} \\ $\longrightarrow$ \texttt{$\{$(E-Type:T, $R_n$, \_:$A_n$)$\}$}\end{tabular}  
        \\
        
    \bottomrule
    \end{tabular}
    }
    \caption{\small{Unification of three tasks. $\longrightarrow$ represents the transformation. For NER, \texttt{E} denotes \texttt{Entity}. For RE, \texttt{S}, \texttt{R}, \texttt{O} represents \texttt{Subject}, \texttt{Relation}, \texttt{Object}, respectively. For EE, \texttt{E}, \texttt{T}, \texttt{R}, \texttt{A} represents \texttt{Event}, \texttt{Trigger}, \texttt{Role}, \texttt{Argument}, respectively while \texttt{n} denotes the number of arguments. \texttt{\_Type} denotes the type. \texttt{\_} represents a pseudo token.}}
    
    \label{tab:scheme}
    \vspace{-5mm}
\end{table}

The schemes of the three tasks can be summarized in Tab.~\ref{tab:scheme}. Through the above descriptions, transformation rules among tasks are observed as follows. RE scheme remains unchanged.
The NER scheme can be attributed to the entity part of the RE scheme.
For EE, if we regard \texttt{Trigger} as \texttt{Subject}, \texttt{Argument} as \texttt{Object}, and \texttt{Role} as \texttt{Relation}, we find it easy to decompose the EE structure into a combination of multiple RE triples. With the transformation rules, we leverage RE as the center, transform NER and EE schemes and thus unify the three IE schemes. 


\paragraph{Integration}\label{sec: unify and integrate}


We uniformly model the scheme of the three IE tasks and integrate the schemes into a series of triples \texttt{(S-Type:S,R,O-Type:O)}. To design an annotation format to fit the unified scheme, we design \texttt{Entity} annotation mode and \texttt{Relation} annotation mode. \texttt{Entity} mode considers entities in NER, subjects, and objects in RE, arguments, and triggers in EE. Similarly, \texttt{Relation} mode considers relations in RE and roles in EE.
It is worth noting that in this way, KG and EKG construction are also integrated besides of IE, NER and EE. 

\subsection{Implementation of Human-Machine-Cooperation}\label{sec:annotate Theory}
\model is designed to support manual labeling and automatic labeling simultaneously.

\paragraph{Manual Labeling}\label{sec: manual}

\begin{table}[!htb]
    \centering
    \resizebox{0.99\columnwidth}{!}{
        \begin{tabular}{cc}
        \toprule
             \textbf{Attr.} & \textbf{Description} \\
        \midrule
            \texttt{isEntity} & Entity or relation \\
            \texttt{suggested} & Markup state (suggested or not) \\
            \texttt{\_id} & Identifier for current markup \\
            \texttt{name} & Name for current entity or relation type \\
            \texttt{labelId} & Id for current entity or relation type \\
            \texttt{source} & Id for subject associated with current relation \\
            \texttt{target} & Id for object associated with current relation \\
            \texttt{start} & Start position for current entity \\
            \texttt{end} & End position for current entity \\
            \texttt{entityText} & Span for current entity \\
        \bottomrule
        \end{tabular}
    }
    
    \caption{Attributes in the \texttt{markup}. \textbf{Attr.}: Attribute. For \texttt{suggested}, \texttt{True} denotes \texttt{suggested} while \texttt{False} denotes \texttt{accepted}.}
    \label{tab:attribute}
\end{table}

We use a dictionary structure called ``markup''~\citep{bikaun2022quickgraph} to record and manage tags. We divide makeups into entity markups and relation markups. The attribute \texttt{isEntity} is used to distinguish them. Moreover, there are many other attributes that are classified as common attributes or private attributes as shown in Tab.~\ref{tab:attribute}. Specially, common attributes mainly include \texttt{suggested}, \texttt{\_id}, \texttt{name}, \texttt{labelId} while private attributes include \texttt{source}, \texttt{target}, \texttt{start}, \texttt{end}, \texttt{entityText}, \etc. 

We follow the standard labeling~\citep{suchanek2007yago,chen2020knowledge}. The process of labeling is to add structured information to the existing graph by linking entities and relationships with relevant types. It involves the following steps: defining the schema, identifying entities, creating nodes, identifying relations, creating edges, validating and refining, iterating and improving. 

\paragraph{Automatic Labeling}\label{sec:auto_labeling}

We adopt ChatIE\footnote{\url{https://github.com/cocacola-lab/ChatIE}}~\citep{wei2023zero}, a SOTA approach for zero-shot information extraction based on prompting ChatGPT. 
It is training-free and suitable for low-resource scenarios. Most importantly, ChatIE is flexible for unification because the type list allows customization.
Given the above advantages, we use ChatIE as the backbone of our automatic annotation algorithm. Then, we make a slight modification by adding the trigger-related prompt template so that it can extract trigger words (Appendix \ref{sec:auto_labeling_all}).

\begin{figure*}[!htp]
\centering
\includegraphics[width=1\linewidth]{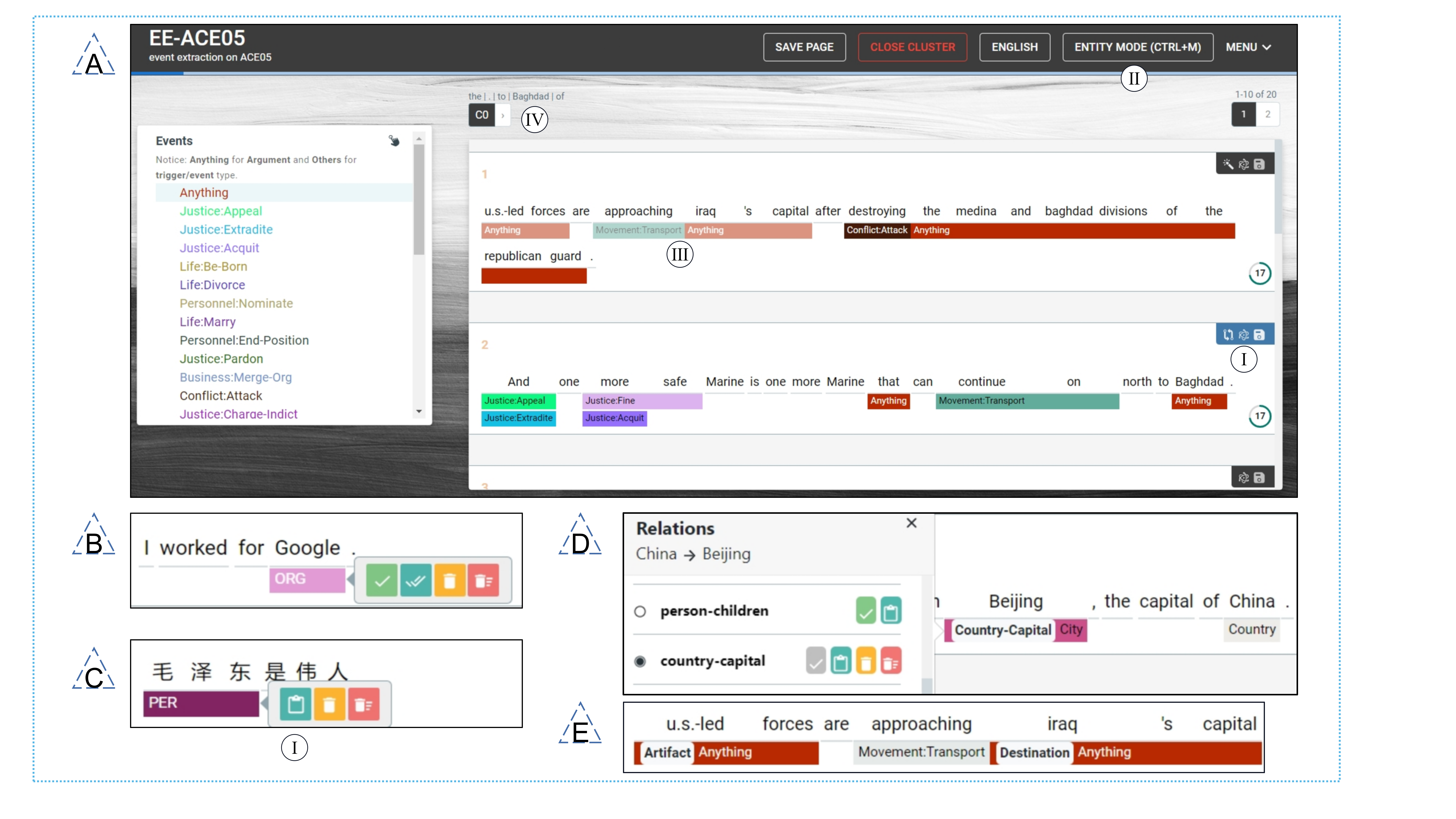}  
\caption{Main Interface of Annotation.}
\label{fig:mainui}
\vspace{-5mm}
\end{figure*}

\subsection{Implementation of Learnability}\label{sec:learnability}

As shown in Fig.~\ref{fig:interaction}, we present the bi-directional interaction workflow between manual and automatic labeling.
\paragraph{Auto $\rightarrow$ Manual} Users can click a specific button to make the system perform automatic labeling.
\model first collects current information to form the input dict, including task, text, and language. Then, the prefix generator searches the knowledge base (which stores high-frequency manual markups) with the current text to generate a prefix prompt. The prefix prompt is fed in the model to enhance domain knowledge, such as \textit{Note: the type of Google is ORG; the relation between James and Google is person-company;...}. Finally, the markup generator generates the final prompt, predicts results, and converts them to markups for human reference. After these steps, the attribute of these markups is set to the suggested state, indicating that these labels are in a pending state and will transition from the suggested state to the accepted state or be deleted after being checked by humans. 
Users click the \texttt{Accept} or \texttt{Delete} button during checking. By utilizing this workflow, users can refer to the automatic annotation results as a reference.

\paragraph{Auto $\leftarrow$ Manual} \model deploys an self-renewal function. It processes high-frequency and informative manual markups and converts them to specific schemes. Then, it updates \model by adding these specific schemes into the knowledge base, which will be used to build a more powerful prefix generator in later iterations. We refer to this process as ``learnability''.

This learnability brings many benefits, such as identifying emerging concepts, domain-specific terms, ambiguous words, \etc. For example, give a text from an electronic product report ``\textit{The middle class likes using Apple.}'', the model may not recognize \textit{Apple} as \texttt{ORG}.
But if humans manually mark \textit{Apple} as \texttt{ORG} in another sentence ``\textit{New Yorker really like Apple phones.}'' and \model perform model updating.
Then the prefix prompt ``\textit{Note: Apple is ORG in ...;}'' will make the model successfully recognize it.

\begin{table*}[!htb]
\centering
\resizebox{0.98\textwidth}{!}{
\begin{tabular}{lccccc|ccccc|ccccc}
\toprule
 & \multicolumn{5}{c}{\textbf{NER}}   & \multicolumn{5}{c}{\textbf{RE}}     & \multicolumn{5}{c}{\textbf{EE}}  \\  
\cline{2-16} 
 & \textit{P}   & \textit{R}    & \textit{F1}      & \textit{Var}   & \textit{Time} & \textit{P}  & \textit{R}    & \textit{F1}  & \textit{Var}   & \textit{Time}  & \textit{P}   & \textit{R}    & \textit{F1}    & \textit{Var}   & \textit{Time}  \\ \midrule
 BRAT   &  -  &   -   &  -   &  -  & 00:42:34   &  - &  - &  - &  - & 01:46:50 & - &  - &  - &  - &   01:34:36 \\ 
\textbf{\model (Auto)}   &  \textbf{86.3}  &   65.1   &  74.2   &  -  & -   &  \textbf{85.5} &  51.5 &  64.2 &  - & - & \textbf{44.2}/82.4 &  38.2/68.9 &  41.0/75.0 &  - &   - \\ 
\textbf{\model (Human)}   &  64.2  &   62.0   &  63.0   &  0.51  &  00:41:27  &  48.1 &  44.2 &  45.8 &  0.69 & 01:27:17 & 36.0/70.5 &  27.5/47.5 &  31.1/56.7 &  0.79/0.53 &   01:12:49 \\ 
\textbf{\model}   &  81.7  &   \textbf{79.3}   &  \textbf{80.4}   &  \textbf{0.39}  &  \textbf{00:40:42}  &  70.8 &  \textbf{71.9} &  \textbf{70.9} &  \textbf{0.41} & \textbf{01:24:01} & 43.4/\textbf{83.4} &  \textbf{45.5}/\textbf{71.3} &  \textbf{44.2}/\textbf{76.9} &  \textbf{0.37}/\textbf{0.09} &   \textbf{01:11:21} \\ \bottomrule
\end{tabular}
}
\caption{Results on three tasks. For EE, the left/right numbers represent Arg-C/Trig-C, respectively. Refer to Appendix. \ref{sec:appendix detailed result} for detailed results.}
\label{table:main_result}
\vspace{-5mm}
\end{table*}

\section{Toolkit Usage}\label{sec:usage}




Core functions and the corresponding implementation make \model powerful, highly productive, and easy to use. We show the toolkit usage by introducing the corresponding UI of three main modules, namely the project creation module, annotation module, and display module in Fig. \ref{fig:mainui}.

\subsection{Project Creation}\label{sec:pc}
\model asks users to first create an account and then start creating a project. The creation process is clear, and the operation is user-friendly and sufficiently guided as shown in Appendix~\ref{sec:appendix p}. 

\subsection{Annotation}\label{sec:an}

\paragraph{Entity/Relation Mode} 
There are two annotation modes, namely entity and relation modes.
In entity mode, the user selects the span and clicks the corresponding type to complete the annotation (Fig. \ref{fig:mainui} B-C).
In relation mode, the user clicks on the subject and then selects the relation type associated with the corresponding object (Fig. \ref{fig:mainui} D-E). 
Moreover, switching between the two modes requires only one click on the toggle (Fig. \ref{fig:mainui} A.II).

\paragraph{Markup State}
There are two markup states, namely accepted state and suggested state. Switching between the two modes requires only one click.
The accepted status indicates that the current label is approved by the human annotator, while the suggested status indicates that it is pending confirmation. Phrases with the suggested state can only be obtained in two ways, \ie, automatic labeling and label propagation.
Phrases marked with the suggested status are highlighted in semitransparent color (Fig. \ref{fig:mainui} A.III) to distinguish themselves from the accepted state.

\paragraph{Automatic Labeling} By clicking the corresponding button (Fig. \ref{fig:mainui} A.I), the predicted results of the machine (\ie, ChatIE) will be transformed into markups to be displayed on the annotation interface.

\paragraph{Label Propagation}
Through propagation, hundreds of entities or relations can be suggested in a single click (Fig. \ref{fig:mainui} C.I).

\paragraph{Annotation Tooltip} Operation icons in the tooltips include \textit{apply one}, \textit{apply all}, \textit{delete one}, \textit{delete all}, \textit{propagate} (Fig. \ref{fig:mainui} B-D).
It is worth noting that \textit{apply} means providing a label as well as changing the label state from the suggested state to the accepted state.

\subsection{Display}\label{sec:dy}
\model offers several features to enhance the user-friendly display. It offers features such as annotation progress display, KG and EKG visualization, double-checking function, and filtering and exporting functions including saving, loading, and quality filtering. See Appendix. \ref{sec:appendix display} for details.

\section{Human Evaluation}\label{sec:experiments}


\subsection{Evaluation Method}\label{subsec:method}
We designed a series of comparison experiments on three IE tasks. We randomly sampled 50 instances from the conllpp~\citep{wang2019crossweigh}, NYT11-HRL~\citep{takanobu2019hierarchical}, and ACE05~\footnote{\url{https://catalog.ldc.upenn.edu/LDC2006T06}} English datasets for the NER, RE, and EE tasks, respectively. For each task, we hired ten humans and randomly divided them equally into the control group (without the automatic labeling module) and the experimental group (with the automatic labeling module).
During the annotation, we record the annotation time, perform performance evaluation, as well as calculate the intra-group variance~\citep{pang-etal-2020-towards} to assess effectiveness, efficiency, and variance.

\emph{Participants:} All participants were senior students (female 4-6 per task) from universities who had passed the qualifications or evidence of English language ability (\eg, CET-6). Every participant was
paid a wage of \$14.17/h.

\emph{Metrics:} Following previous work~\citep{wei2023zero}, we adopt Micro F1 as the evaluation metric.
For NER, the predicted entity is correct only if its whole span and type are correct.
For RE, an extracted triple is considered as correct if the whole span of both head and tail entities, as well as the relation, are all correct.
For EE~\citep{lin-etal-2020-joint}, an argument is correctly identified only if its whole span, role label, and event type match the ground truth (Arg-C).
A trigger is correctly identified only if its whole span and event type match the golden trigger (Trig-C).
We calculate the intra-group variance as follows: $var = 1 - \frac{1}{|D|\cdot C_{|G|}^2}\sum\sum_{i<j\in G} get\_f1\_list(g_i, g_j)$, where $G$ denotes the experimental or control group, $g$ denotes the sequence of annotation results and $D$ denotes the dataset.

\emph{Procedure:} Participants signed an informed consent form before being assigned accounts to annotate with/without the automatic labeling module. Time spent on annotation was recorded for evaluation and statistics. For more information on the human evaluation procedure, please refer to Appendix. \ref{sec:appendix human procedure}.

\subsection{Results}\label{subsec:result}
The results for three tasks are shown in Tab. \ref{table:main_result}. \model (\textbf{Auto}) denotes pure automatic labeling. \model (\textbf{Human}) denotes humans using \model without assistance from automatic labeling. \model denotes complete human-machine cooperation. Results show that \model significantly improves annotation quality, efficiency, and stability.

On NER, RE, and EE tasks, the average improvement \wrt effectiveness is 18.75\%. The average improvement \wrt variance is 0.315. Time is most affected by external or human factors, but nonetheless, speedups of 1.8\%, 3.9\%, and 2.1\% were achieved on NER, RE, and EE tasks, respectively. In addition, compared to BRAT, the annotation speed of CollabKG exceeds 4.6\%, 27.2\%, and 32.6\% for NER, RE and EE tasks, respectively.




\section{Conclusion}\label{sec:conclusion}
We present \model an open-source IE annotation tool for KG and EKG construction.
Ultimately, we conducted an extensive human evaluation to quantitatively demonstrate that \model can significantly improve annotation quality, efficiency, and stability as well as qualitative comparisons with other existing open-source tools.

\section*{Limitations}
The complexity of the IE annotation tasks may require significant human expertise and may lead to errors or inconsistencies in the annotations. Our human-machine-cooperative \model helps in terms of reducing labor and eliminating inconsistencies. However, the cost of developing and maintaining the LLM as SaaS (Software as a Service) may be high, particularly if the LLM requires long input while updates are accumulated. We require a lightweight LLM.

\section*{Ethics Statement}
All participants in this study will be fully informed of the nature of the study and will be required to provide informed consent prior to participation. All personal data will be kept confidential and anonymous.


\bibliography{anthology,emnlp2023}

\begin{thebibliography}{40}
\expandafter\ifx\csname natexlab\endcsname\relax\def\natexlab#1{#1}\fi

\bibitem[{Abrami et~al.(2019)Abrami, Mehler, Lücking, Rieb, and
  Helfrich}]{inproceedings}
Giuseppe Abrami, Alexander Mehler, Andy Lücking, Elias Rieb, and Philipp
  Helfrich. 2019.
\newblock Textannotator: A flexible framework for semantic annotations.

\bibitem[{Akbik et~al.(2019)Akbik, Bergmann, Blythe, Rasul, Schweter, and
  Vollgraf}]{akbik-etal-2019-flair}
Alan Akbik, Tanja Bergmann, Duncan Blythe, Kashif Rasul, Stefan Schweter, and
  Roland Vollgraf. 2019.
\newblock \href {https://doi.org/10.18653/v1/N19-4010} {{FLAIR}: An easy-to-use
  framework for state-of-the-art {NLP}}.
\newblock In \emph{Proceedings of the 2019 Conference of the North {A}merican
  Chapter of the Association for Computational Linguistics (Demonstrations)},
  pages 54--59, Minneapolis, Minnesota. Association for Computational
  Linguistics.

\bibitem[{Aone and Ramos-Santacruz(2000)}]{aone-ramos-santacruz-2000-rees}
Chinatsu Aone and Mila Ramos-Santacruz. 2000.
\newblock \href {https://doi.org/10.3115/974147.974158} {{REES}: A large-scale
  relation and event extraction system}.
\newblock In \emph{Sixth Applied Natural Language Processing Conference}, pages
  76--83, Seattle, Washington, USA. Association for Computational Linguistics.

\bibitem[{Bao et~al.(2016)Bao, Duan, Yan, Zhou, and Zhao}]{bao2016constraint}
Junwei Bao, Nan Duan, Zhao Yan, Ming Zhou, and Tiejun Zhao. 2016.
\newblock Constraint-based question answering with knowledge graph.
\newblock In \emph{Proceedings of COLING 2016, the 26th international
  conference on computational linguistics: technical papers}, pages 2503--2514.

\bibitem[{Bien et~al.(2018)Bien, Rajpurkar, Ball, Irvin, Park, Jones, Bereket,
  Patel, Yeom, Shpanskaya et~al.}]{bien2018deep}
Nicholas Bien, Pranav Rajpurkar, Robyn~L Ball, Jeremy Irvin, Allison Park, Erik
  Jones, Michael Bereket, Bhavik~N Patel, Kristen~W Yeom, Katie Shpanskaya,
  et~al. 2018.
\newblock Deep-learning-assisted diagnosis for knee magnetic resonance imaging:
  development and retrospective validation of mrnet.
\newblock \emph{PLoS medicine}, 15(11):e1002699.

\bibitem[{Bikaun et~al.(2022)Bikaun, Stewart, and Liu}]{bikaun2022quickgraph}
Tyler Bikaun, Michael Stewart, and Wei Liu. 2022.
\newblock Quickgraph: A rapid annotation tool for knowledge graph extraction
  from technical text.
\newblock In \emph{Proceedings of the 60th Annual Meeting of the Association
  for Computational Linguistics: System Demonstrations}, pages 270--278.

\bibitem[{Chen et~al.(2020)Chen, Wang, Zhao, Cheng, Zhao, and
  Duan}]{chen2020knowledge}
Zhe Chen, Yuehan Wang, Bin Zhao, Jing Cheng, Xin Zhao, and Zongtao Duan. 2020.
\newblock Knowledge graph completion: A review.
\newblock \emph{Ieee Access}, 8:192435--192456.

\bibitem[{Guan et~al.(2022)Guan, Cheng, Bai, Zhang, Li, Zeng, Jin, and
  Guo}]{guan2022event}
Saiping Guan, Xueqi Cheng, Long Bai, Fujun Zhang, Zixuan Li, Yutao Zeng,
  Xiaolong Jin, and Jiafeng Guo. 2022.
\newblock What is event knowledge graph: a survey.
\newblock \emph{IEEE Transactions on Knowledge and Data Engineering}.

\bibitem[{Han et~al.(2019)Han, Gao, Yao, Ye, Liu, and
  Sun}]{han-etal-2019-opennre}
Xu~Han, Tianyu Gao, Yuan Yao, Deming Ye, Zhiyuan Liu, and Maosong Sun. 2019.
\newblock \href {https://doi.org/10.18653/v1/D19-3029} {{O}pen{NRE}: An open
  and extensible toolkit for neural relation extraction}.
\newblock In \emph{Proceedings of the 2019 Conference on Empirical Methods in
  Natural Language Processing and the 9th International Joint Conference on
  Natural Language Processing (EMNLP-IJCNLP): System Demonstrations}, pages
  169--174, Hong Kong, China. Association for Computational Linguistics.

\bibitem[{Islamaj et~al.(2020)Islamaj, Kwon, Kim, and Lu}]{10.1093/nar/gkaa333}
Rezarta Islamaj, Dongseop Kwon, Sun Kim, and Zhiyong Lu. 2020.
\newblock \href {https://doi.org/10.1093/nar/gkaa333} {{TeamTat: a
  collaborative text annotation tool}}.
\newblock \emph{Nucleic Acids Research}, 48(W1):W5--W11.

\bibitem[{Jin et~al.(2021)Jin, Chen, Sui, Wang, Xue, and Zhao}]{jin2021cogie}
Zhuoran Jin, Yubo Chen, Dianbo Sui, Chenhao Wang, Zhipeng Xue, and Jun Zhao.
  2021.
\newblock Cogie: An information extraction toolkit for bridging texts and
  cognet.
\newblock In \emph{Proceedings of the 59th Annual Meeting of the Association
  for Computational Linguistics and the 11th International Joint Conference on
  Natural Language Processing: System Demonstrations}, pages 92--98.

\bibitem[{Kejriwal(2022)}]{kejriwal2022knowledge}
Mayank Kejriwal. 2022.
\newblock Knowledge graphs: Constructing, completing, and effectively applying
  knowledge graphs in tourism.
\newblock In \emph{Applied Data Science in Tourism: Interdisciplinary
  Approaches, Methodologies, and Applications}, pages 423--449. Springer.

\bibitem[{Klie et~al.(2018)Klie, Bugert, Boullosa, Eckart~de Castilho, and
  Gurevych}]{klie-etal-2018-inception}
Jan-Christoph Klie, Michael Bugert, Beto Boullosa, Richard Eckart~de Castilho,
  and Iryna Gurevych. 2018.
\newblock \href {https://aclanthology.org/C18-2002} {The {INCE}p{TION}
  platform: Machine-assisted and knowledge-oriented interactive annotation}.
\newblock In \emph{Proceedings of the 27th International Conference on
  Computational Linguistics: System Demonstrations}, pages 5--9, Santa Fe, New
  Mexico. Association for Computational Linguistics.

\bibitem[{Kummerfeld(2019)}]{kummerfeld-2019-slate}
Jonathan~K. Kummerfeld. 2019.
\newblock \href {https://doi.org/10.18653/v1/P19-3002} {{SLATE}: A
  super-lightweight annotation tool for experts}.
\newblock In \emph{Proceedings of the 57th Annual Meeting of the Association
  for Computational Linguistics: System Demonstrations}, pages 7--12, Florence,
  Italy. Association for Computational Linguistics.

\bibitem[{Li et~al.(2020)Li, Wang, Yan, Wang, Li, Jiang, Sun, Tang, Chang, Wang
  et~al.}]{li2020real}
Linfeng Li, Peng Wang, Jun Yan, Yao Wang, Simin Li, Jinpeng Jiang, Zhe Sun,
  Buzhou Tang, Tsung-Hui Chang, Shenghui Wang, et~al. 2020.
\newblock Real-world data medical knowledge graph: construction and
  applications.
\newblock \emph{Artificial intelligence in medicine}, 103:101817.

\bibitem[{Li et~al.(2021)Li, Yu, Quangang, and Liu}]{li-etal-2021-fitannotator}
Yanzeng Li, Bowen Yu, Li~Quangang, and Tingwen Liu. 2021.
\newblock \href {https://doi.org/10.18653/v1/2021.naacl-demos.5}
  {{FITA}nnotator: A flexible and intelligent text annotation system}.
\newblock In \emph{Proceedings of the 2021 Conference of the North American
  Chapter of the Association for Computational Linguistics: Human Language
  Technologies: Demonstrations}, pages 35--41, Online. Association for
  Computational Linguistics.

\bibitem[{Lin et~al.(2019)Lin, Chen, Chen, and Ren}]{lin2019kagnet}
Bill~Yuchen Lin, Xinyue Chen, Jamin Chen, and Xiang Ren. 2019.
\newblock Kagnet: Knowledge-aware graph networks for commonsense reasoning.
\newblock \emph{arXiv preprint arXiv:1909.02151}.

\bibitem[{Lin et~al.(2020)Lin, Ji, Huang, and Wu}]{lin-etal-2020-joint}
Ying Lin, Heng Ji, Fei Huang, and Lingfei Wu. 2020.
\newblock \href {https://doi.org/10.18653/v1/2020.acl-main.713} {A joint neural
  model for information extraction with global features}.
\newblock In \emph{Proceedings of the 58th Annual Meeting of the Association
  for Computational Linguistics}, pages 7999--8009, Online. Association for
  Computational Linguistics.

\bibitem[{Luan et~al.(2018)Luan, He, Ostendorf, and Hajishirzi}]{luan2018multi}
Yi~Luan, Luheng He, Mari Ostendorf, and Hannaneh Hajishirzi. 2018.
\newblock Multi-task identification of entities, relations, and coreference for
  scientific knowledge graph construction.
\newblock \emph{arXiv preprint arXiv:1808.09602}.

\bibitem[{Neves and Ševa(2019)}]{10.1093/bib/bbz130}
Mariana Neves and Jurica Ševa. 2019.
\newblock \href {https://doi.org/10.1093/bib/bbz130} {{An extensive review of
  tools for manual annotation of documents}}.
\newblock \emph{Briefings in Bioinformatics}, 22(1):146--163.

\bibitem[{Nghiem and Ananiadou(2018)}]{nghiem-ananiadou-2018-aplenty}
Minh-Quoc Nghiem and Sophia Ananiadou. 2018.
\newblock \href {https://doi.org/10.18653/v1/D18-2019} {{APL}enty: annotation
  tool for creating high-quality datasets using active and proactive learning}.
\newblock In \emph{Proceedings of the 2018 Conference on Empirical Methods in
  Natural Language Processing: System Demonstrations}, pages 108--113,
  Brussels, Belgium. Association for Computational Linguistics.

\bibitem[{Pang et~al.(2020)Pang, Nijkamp, Han, Zhou, Liu, and
  Tu}]{pang-etal-2020-towards}
Bo~Pang, Erik Nijkamp, Wenjuan Han, Linqi Zhou, Yixian Liu, and Kewei Tu. 2020.
\newblock \href {https://doi.org/10.18653/v1/2020.acl-main.333} {Towards
  holistic and automatic evaluation of open-domain dialogue generation}.
\newblock In \emph{Proceedings of the 58th Annual Meeting of the Association
  for Computational Linguistics}, pages 3619--3629, Online. Association for
  Computational Linguistics.

\bibitem[{Reimers and Gurevych(2019)}]{reimers-gurevych-2019-sentence}
Nils Reimers and Iryna Gurevych. 2019.
\newblock \href {https://doi.org/10.18653/v1/D19-1410} {Sentence-{BERT}:
  Sentence embeddings using {S}iamese {BERT}-networks}.
\newblock In \emph{Proceedings of the 2019 Conference on Empirical Methods in
  Natural Language Processing and the 9th International Joint Conference on
  Natural Language Processing (EMNLP-IJCNLP)}, pages 3982--3992, Hong Kong,
  China. Association for Computational Linguistics.

\bibitem[{Souza~Costa et~al.(2020)Souza~Costa, Gottschalk, and
  Demidova}]{souza2020event}
Tarc{\'\i}sio Souza~Costa, Simon Gottschalk, and Elena Demidova. 2020.
\newblock Event-qa: A dataset for event-centric question answering over
  knowledge graphs.
\newblock In \emph{Proceedings of the 29th ACM international conference on
  information \& knowledge management}, pages 3157--3164.

\bibitem[{Stenetorp et~al.(2012)Stenetorp, Pyysalo, Topi{\'c}, Ohta, Ananiadou,
  and Tsujii}]{stenetorp-etal-2012-brat}
Pontus Stenetorp, Sampo Pyysalo, Goran Topi{\'c}, Tomoko Ohta, Sophia
  Ananiadou, and Jun{'}ichi Tsujii. 2012.
\newblock \href {https://aclanthology.org/E12-2021} {brat: a web-based tool for
  {NLP}-assisted text annotation}.
\newblock In \emph{Proceedings of the Demonstrations at the 13th Conference of
  the {E}uropean Chapter of the Association for Computational Linguistics},
  pages 102--107, Avignon, France. Association for Computational Linguistics.

\bibitem[{Stewart et~al.(2019)Stewart, Liu, and
  Cardell-Oliver}]{stewart-etal-2019-redcoat}
Michael Stewart, Wei Liu, and Rachel Cardell-Oliver. 2019.
\newblock \href {https://doi.org/10.18653/v1/D19-3033} {{R}edcoat: A
  collaborative annotation tool for hierarchical entity typing}.
\newblock In \emph{Proceedings of the 2019 Conference on Empirical Methods in
  Natural Language Processing and the 9th International Joint Conference on
  Natural Language Processing (EMNLP-IJCNLP): System Demonstrations}, pages
  193--198, Hong Kong, China. Association for Computational Linguistics.

\bibitem[{Suchanek et~al.(2007)Suchanek, Kasneci, and
  Weikum}]{suchanek2007yago}
Fabian~M Suchanek, Gjergji Kasneci, and Gerhard Weikum. 2007.
\newblock Yago: a core of semantic knowledge.
\newblock In \emph{Proceedings of the 16th international conference on World
  Wide Web}, pages 697--706.

\bibitem[{Takanobu et~al.(2019)Takanobu, Zhang, Liu, and
  Huang}]{takanobu2019hierarchical}
Ryuichi Takanobu, Tianyang Zhang, Jiexi Liu, and Minlie Huang. 2019.
\newblock A hierarchical framework for relation extraction with reinforcement
  learning.
\newblock In \emph{Proceedings of the AAAI conference on artificial
  intelligence}, volume~33, pages 7072--7079.

\bibitem[{Tang et~al.(2020)Tang, Su, Chen, Qu, and Ding}]{9378107}
Mingwei Tang, Cui Su, Haihua Chen, Jingye Qu, and Junhua Ding. 2020.
\newblock \href {https://doi.org/10.1109/BigData50022.2020.9378107} {Salkg: A
  semantic annotation system for building a high-quality legal knowledge
  graph}.
\newblock In \emph{2020 IEEE International Conference on Big Data (Big Data)},
  pages 2153--2159.

\bibitem[{Valenzuela-Esc{\'a}rcega et~al.(2015)Valenzuela-Esc{\'a}rcega,
  Hahn-Powell, Surdeanu, and Hicks}]{valenzuela-escarcega-etal-2015-domain}
Marco~A. Valenzuela-Esc{\'a}rcega, Gus Hahn-Powell, Mihai Surdeanu, and Thomas
  Hicks. 2015.
\newblock \href {https://doi.org/10.3115/v1/P15-4022} {A domain-independent
  rule-based framework for event extraction}.
\newblock In \emph{Proceedings of {ACL}-{IJCNLP} 2015 System Demonstrations},
  pages 127--132, Beijing, China. Association for Computational Linguistics and
  The Asian Federation of Natural Language Processing.

\bibitem[{Wang et~al.(2019{\natexlab{a}})Wang, He, Cao, Liu, and
  Chua}]{wang2019kgat}
Xiang Wang, Xiangnan He, Yixin Cao, Meng Liu, and Tat-Seng Chua.
  2019{\natexlab{a}}.
\newblock Kgat: Knowledge graph attention network for recommendation.
\newblock In \emph{Proceedings of the 25th ACM SIGKDD international conference
  on knowledge discovery \& data mining}, pages 950--958.

\bibitem[{Wang et~al.(2019{\natexlab{b}})Wang, Shang, Liu, Lu, Liu, and
  Han}]{wang2019crossweigh}
Zihan Wang, Jingbo Shang, Liyuan Liu, Lihao Lu, Jiacheng Liu, and Jiawei Han.
  2019{\natexlab{b}}.
\newblock Crossweigh: Training named entity tagger from imperfect annotations.
\newblock \emph{arXiv preprint arXiv:1909.01441}.

\bibitem[{Wei et~al.(2023)Wei, Cui, Cheng, Wang, Zhang, Huang, Xie, Xu, Chen,
  Zhang et~al.}]{wei2023zero}
Xiang Wei, Xingyu Cui, Ning Cheng, Xiaobin Wang, Xin Zhang, Shen Huang, Pengjun
  Xie, Jinan Xu, Yufeng Chen, Meishan Zhang, et~al. 2023.
\newblock Zero-shot information extraction via chatting with chatgpt.
\newblock \emph{arXiv preprint arXiv:2302.10205}.

\bibitem[{Wei et~al.(2019)Wei, Su, Wang, Tian, and Chang}]{wei2019novel}
Zhepei Wei, Jianlin Su, Yue Wang, Yuan Tian, and Yi~Chang. 2019.
\newblock A novel cascade binary tagging framework for relational triple
  extraction.
\newblock \emph{arXiv preprint arXiv:1909.03227}.

\bibitem[{Wen et~al.(2021)Wen, Lin, Lai, Pan, Li, Lin, Zhou, Li, Wang, Zhang,
  Yu, Dong, Wang, Fung, Mishra, Lyu, Sur{\'\i}s, Chen, Brown, Palmer,
  Callison-Burch, Vondrick, Han, Roth, Chang, and Ji}]{wen-etal-2021-resin}
Haoyang Wen, Ying Lin, Tuan Lai, Xiaoman Pan, Sha Li, Xudong Lin, Ben Zhou,
  Manling Li, Haoyu Wang, Hongming Zhang, Xiaodong Yu, Alexander Dong,
  Zhenhailong Wang, Yi~Fung, Piyush Mishra, Qing Lyu, D{\'\i}dac Sur{\'\i}s,
  Brian Chen, Susan~Windisch Brown, Martha Palmer, Chris Callison-Burch, Carl
  Vondrick, Jiawei Han, Dan Roth, Shih-Fu Chang, and Heng Ji. 2021.
\newblock \href {https://doi.org/10.18653/v1/2021.naacl-demos.16} {{RESIN}: A
  dockerized schema-guided cross-document cross-lingual cross-media information
  extraction and event tracking system}.
\newblock In \emph{Proceedings of the 2021 Conference of the North American
  Chapter of the Association for Computational Linguistics: Human Language
  Technologies: Demonstrations}, pages 133--143, Online. Association for
  Computational Linguistics.

\bibitem[{Yang et~al.(2020)Yang, Li, Zhang, Zhang, and Qi}]{yang2020temporal}
Chengbiao Yang, Weizhuo Li, Xiaoping Zhang, Runshun Zhang, and Guilin Qi. 2020.
\newblock A temporal semantic search system for traditional chinese medicine
  based on temporal knowledge graphs.
\newblock In \emph{Semantic Technology: 9th Joint International Conference,
  JIST 2019, Hangzhou, China, November 25--27, 2019, Revised Selected Papers
  9}, pages 13--20. Springer.

\bibitem[{Yimam et~al.(2013)Yimam, Gurevych, Eckart~de Castilho, and
  Biemann}]{yimam-etal-2013-webanno}
Seid~Muhie Yimam, Iryna Gurevych, Richard Eckart~de Castilho, and Chris
  Biemann. 2013.
\newblock \href {https://aclanthology.org/P13-4001} {{W}eb{A}nno: A flexible,
  web-based and visually supported system for distributed annotations}.
\newblock In \emph{Proceedings of the 51st Annual Meeting of the Association
  for Computational Linguistics: System Demonstrations}, pages 1--6, Sofia,
  Bulgaria. Association for Computational Linguistics.

\bibitem[{Zhang et~al.(2022)Zhang, Xu, Tao, Yu, Ye, Xie, Chen, Li, Li, Liang
  et~al.}]{zhang2022deepke}
Ningyu Zhang, Xin Xu, Liankuan Tao, Haiyang Yu, Hongbin Ye, Xin Xie, Xiang
  Chen, Zhoubo Li, Lei Li, Xiaozhuan Liang, et~al. 2022.
\newblock Deepke: A deep learning based knowledge extraction toolkit for
  knowledge base population.
\newblock \emph{arXiv preprint arXiv:2201.03335}.

\bibitem[{Zhao et~al.(2021)Zhao, Chen, Xing, and Miao}]{zhao2021brain}
Xuejiao Zhao, Huanhuan Chen, Zhenchang Xing, and Chunyan Miao. 2021.
\newblock Brain-inspired search engine assistant based on knowledge graph.
\newblock \emph{IEEE Transactions on Neural Networks and Learning Systems}.

\bibitem[{Zou(2020)}]{zou2020survey}
Xiaohan Zou. 2020.
\newblock A survey on application of knowledge graph.
\newblock In \emph{Journal of Physics: Conference Series}, volume 1487, page
  012016. IOP Publishing.

\end{thebibliography}
\bibliographystyle{acl_natbib}

\appendix

\section{System Architecture}
\label{sec:appendix system}
\begin{figure}[htp]
\centering
\includegraphics[width=1\linewidth]{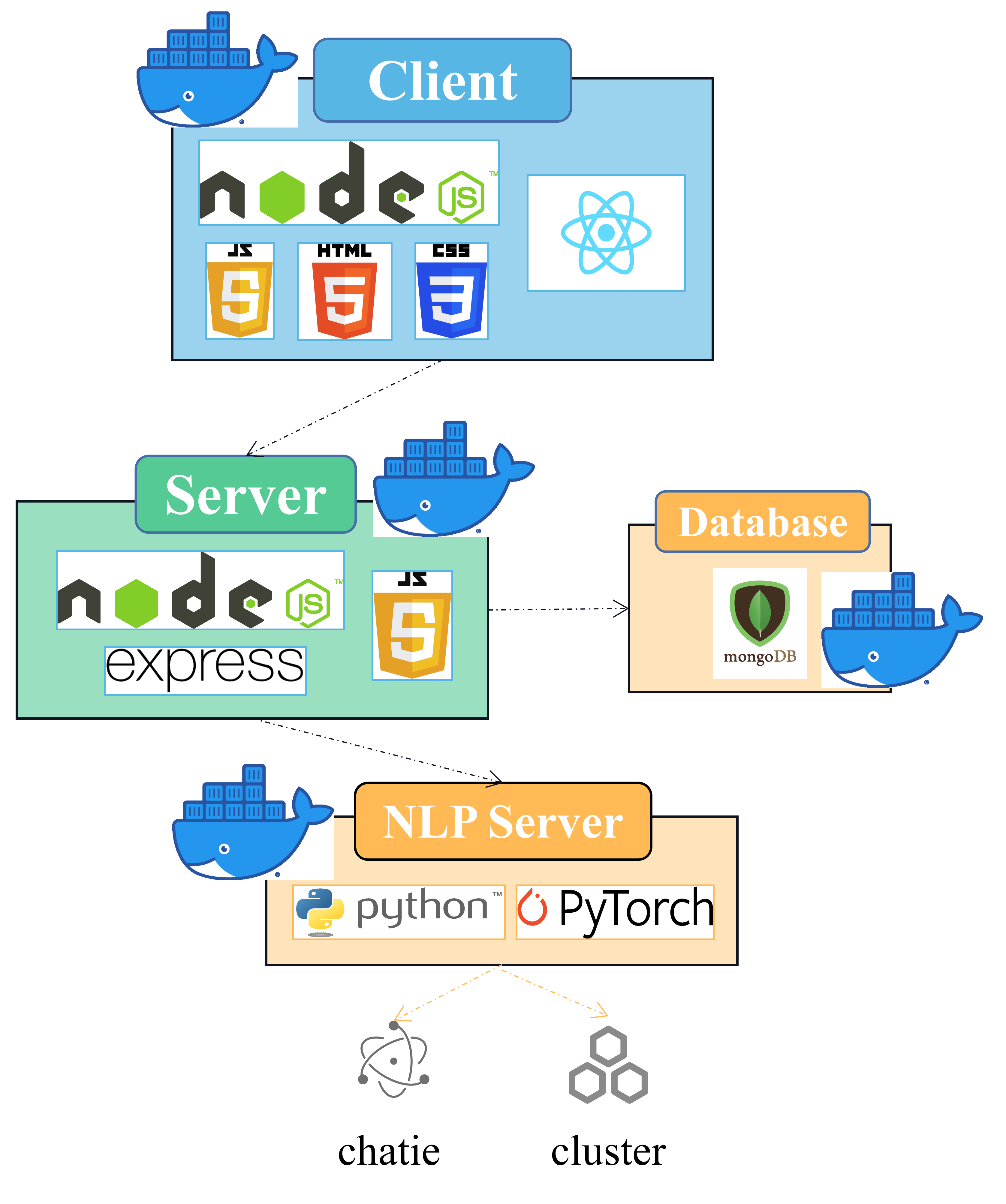}  
\caption{System Architecture.}
\label{fig:system}
\vspace{-0.2em}
\end{figure}

As shown in Fig.~\ref{fig:system}, we use modern full-stack framework MERN~\footnote{MongoDB-Express-React-Node \url{https://www.mongodb.com/mern-stack}}, Docker and Python to build \model.
It consists of four components wrapped in a Docker container, namely web client, server, NLP server, and database.
The NLP server deploys ChatIE to obtain automatic annotation results.
In addition, the database is crucial, storing and managing information such as projects, texts, users, \etc. This is achieved by maintaining three collections including \texttt{Project}, \texttt{Text}, and \texttt{User}.
\texttt{Project} stores the task details of the project, whether to perform the model update, semantic clustering, preprocessing, \etc. \texttt{Text} manages samples, markups, \etc. \texttt{User} stores the information of users like name and password.

\section{Related Works}
\label{sec:appendix related works}
There are many existing open-source IE (\ie., NER, RE, and EE) annotation tools for KG or EKG construction. 
We will describe them from several perspectives such as applicable tasks and annotation styles.

From the perspective of the three applicable tasks, FLAIR~\citep{akbik-etal-2019-flair}, OpenNRE~\citep{han-etal-2019-opennre}, ODIN~\citep{valenzuela-escarcega-etal-2015-domain}, FitAnnotator~\citep{li-etal-2021-fitannotator}, WebAnno~\citep{yimam-etal-2013-webanno}, Redcoat~\citep{stewart-etal-2019-redcoat} and APLenty~\citep{nghiem-ananiadou-2018-aplenty} only focus on a single task. DeepKE~\citep{zhang2022deepke}, Quickgraph~\citep{bikaun2022quickgraph}, INCEpTION~\citep{klie-etal-2018-inception}, TeamTat~\citep{10.1093/nar/gkaa333}, TextAnnotator~\citep{inproceedings}, REES~\citep{aone-ramos-santacruz-2000-rees} and SALKG~\citep{9378107} support two tasks. CogIE~\citep{jin2021cogie}, BRAT~\citep{stenetorp-etal-2012-brat}, SLATE~\citep{kummerfeld-2019-slate} and RESIN~\citep{wen-etal-2021-resin} support all three tasks. However, CogIE and RESIN only support automatic labeling and require training, which is not suitable for low-resource scenarios where data is insufficient for training from scratch.
BRAT and SLATE only support manual labeling. Moreover, BRAT sometimes is criticized for its difficulties in deployment~\citep{10.1093/bib/bbz130}. SLATE is a command-line-based tool so it is not user-friendly.

From the perspective of the annotation styles (\ie, automatic and manual labeling).
INCEpTION, TextAnnotator, FitAnnotator, and APLenty support two annotation styles immediately. However, INCEpTION and TextAnnotator don't support the learnability function (namely, self-renewal). Although FitAnnotaor and APLenty utilize active learning to support two-way interaction, this is inflexible, not real-time, and requires a training process. In addition, as mentioned earlier, they only support a single task.

\section{Detailed Results}
\label{sec:appendix detailed result}
To note, the standard labeling process includes multiple rounds of labeling, including validating and refining, iterating and improving. 
Validating and refining the annotations in the knowledge graph aims to ensure that the labels accurately represent the domain knowledge by double-checking.
Iterating and improving denote continuously iterating and improving the knowledge graph based on feedback from multiple turns.
In this work, all results are obtained in one round of annotation, so the numbers may seem to be low.

\textbf{NER} The results are presented in Tab. \ref{table:ner_result}. The experimental group consisted of No. 6 to 10, while the control group consisted of No. 1 to 5.
Before calculating the metrics we eliminated No.3 and No.8 because of their poor annotating quality.
\begin{table}
\centering
\resizebox{0.70\linewidth}{!}{
\begin{tabular}{lcccc}
\toprule
 & \textbf{P}   & \textbf{R}    & \textbf{F1}   & \textbf{Time}\\ \midrule
\textbf{P1}   &  56.4  &   58.5   &  57.4   & 00:40:24  \\ 
\textbf{P2}   &  70.3  &   60.4   &  65.0   & 00:42:21  \\ 
\textbf{P3}   &  48.1  &   58.5   &  52.8   & 00:41:03  \\ 
\textbf{P4}   &  71.4  &   66.0   &  68.6   & 00:42:07  \\ 
\textbf{P5}   &  58.8  &   63.2   &  60.9   & 00:40:56  \\ 
\textbf{P6}   &  79.4  &   80.2   &  79.8   & 00:40:31  \\ 
\textbf{P7}   &  87.8  &   81.1   &  84.3   & 00:40:32  \\ 
\textbf{P8}   &  73.3  &   72.6   &  73.0   & 00:41:17  \\ 
\textbf{P9}   &  83.8  &   78.3   &  81.0   & 00:40:09  \\ 
\textbf{P10}   &  75.9  &   77.4   &  76.6   & 00:41:40  \\ 

\bottomrule
\end{tabular}
}
\caption{Human evaluation results on NER. \textbf{P} denotes participant.}
\label{table:ner_result}
\end{table}

\textbf{RE} The results are presented in Tab. \ref{table:re_result}. The experimental group consisted of No. 6 to 10.
Before calculating the metrics we eliminated No.5 and No.6 because of their long annotation time or poor annotation quality.
It is worth noting that since NYT11-HRL is obtained by remote supervision, the gold annotation does not cover all entities and relationships~\citep{wei2019novel}. Therefore, we re-examined and relabeled the 50 samples as the gold label.
\begin{table}
\centering
\resizebox{0.7\linewidth}{!}{
\begin{tabular}{lcccc}
\toprule
 & \textbf{P}   & \textbf{R}    & \textbf{F1}   & \textbf{Time}\\ \midrule
\textbf{P1}   &  57.0  &   47.6   &  51.9   & 01:29:42  \\ 
\textbf{P2}   &  50.0  &   50.5   &  50.2   & 01:32:39  \\ 
\textbf{P3}   &  35.3  &   39.8   &  37.4   & 01:23:44  \\ 
\textbf{P4}   &  50.0  &   38.8   &  43.7   & 01:23:01  \\ 
\textbf{P5}   &  47.9  &   33.0   &  39.1   & 02:15:39  \\ 
\textbf{P6}   &  55.4  &   60.2   &  57.7   & 01:21:40  \\ 
\textbf{P7}   &  68.1  &   62.1   &  65.0   & 01:14:04  \\ 
\textbf{P8}   &  80.9  &   73.8   &  77.2   & 01:21:28  \\ 
\textbf{P9}   &  75.3  &   73.8   &  74.5   & 01:23:56  \\ 
\textbf{P10}   &  58.8  &   77.7   &  67.0   & 01:36:39  \\ 

\bottomrule
\end{tabular}
}
\caption{Human evaluation results on RE. \textbf{P} denotes participant.}
\label{table:re_result}
\end{table}

\textbf{EE} The results are presented in Tab. \ref{table:ee_result}. The experimental group consisted of No. 6 to 10.
No. 5 participant was absent for personal reasons, so to align with the control group, we eliminated the results of No. 6 participant (poor annotation quality).
It is worth noting that because there are so many tags in ACE05 (namely, 33), annotating them all would be too tricky.
Therefore, we narrowed the tag range to 12 (within 50 samples).

\begin{table}
\centering
\resizebox{0.9\linewidth}{!}{
\begin{tabular}{lcccc}
\toprule
 & \textbf{P}   & \textbf{R}    & \textbf{F1}   & \textbf{Time}\\ \midrule
\textbf{P1}   &  37.3/69.2  &   25.5/44.3   &  30.3/54.0   & 01:10:44  \\ 
\textbf{P2}   &  43.4/75.0  &   30.0/49.2   &  35.5/59.4   & 01:27:25  \\ 
\textbf{P3}   &  26.5/65.2  &   23.6/49.2   &  25.0/56.1   & 01:05:34  \\ 
\textbf{P4}   &  36.6/72.5  &   30.9/47.6   &  33.5/57.4   & 01:07:34  \\ 
\textbf{P5}   &  -  &   -   &  -   & -  \\ 
\textbf{P6}   &  37.7/84.6  &   41.8/72.1   &  39.7/77.9   & 01:04:51  \\ 
\textbf{P7}   &  47.4/86.0  &   40.9/70.5   &  43.9/77.5   & 01:02:48  \\ 
\textbf{P8}   &  44.9/82.7  &   48.2/70.5   &  46.5/76.1   & 01:33:12  \\ 
\textbf{P9}   &  37.2/78.6  &   46.4/72.1   &  41.3/75.2   & 01:07:02  \\ 
\textbf{P10}   &  44.0/86.3  &   46.4/72.1   &  45.1/78.6   & 01:02:21  \\ 

\bottomrule
\end{tabular}
}
\caption{Human evaluation results on EE. The left and right numbers in Column \textbf{F1} represent Arg-C and Trig-C, respectively. \textbf{P} denotes participant.}
\label{table:ee_result}
\end{table}

\section{Implementation of Unification}\label{sec:implementation_of_unification_all}
\begin{figure*}[htp]
\centering
\includegraphics[width=1\linewidth]{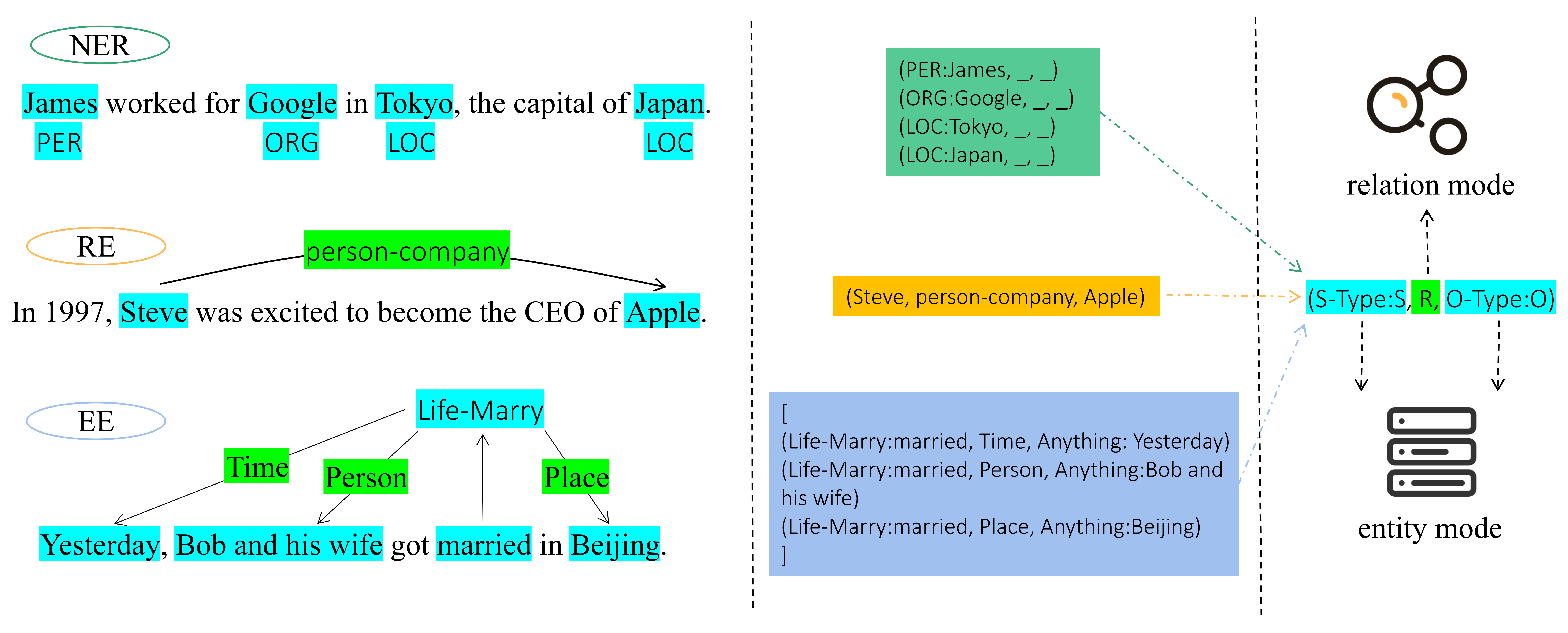}  
\caption{Illustration of unifying NER, IE, and EE tasks.}
\label{fig:unify task}
\end{figure*}

As shown in Fig.~\ref{fig:unify task}, to unify NER, IE, and EE tasks, we first observe their schemes and summarize the transformation rules among them. Then we uniformly model the schemes utilizing these rules and design an annotation format.

\section{Automatic Labeling}\label{sec:auto_labeling_all}
\begin{table}[!htb]
    \centering
    \resizebox{0.95\columnwidth}{!}{
        \begin{tabular}{c|c}
        \toprule
             NER & \texttt{[entity type 1, ..., entity type n]} \\
        \midrule
            RE & \texttt{\{relation type 1: [subject1, object1],...\}} \\
        \midrule
            EE & \texttt{\{event type 1: [argument role 1, ...],...\}} \\
        \bottomrule
        \end{tabular}
    }
    \caption{Type list format for three IE tasks.}
    \label{tab:type_list}
\end{table}


We have adopted ChatIE\footnote{\url{https://github.com/cocacola-lab/ChatIE}}~\citep{wei2023zero} as our approach for zero-shot information extraction, based on ChatGPT. ChatIE has shown impressive performance, even surpassing some full-shot models across various datasets. Its flexibility for customization is also a notable advantage because the type list allows customization (refer to Tab.~\ref{tab:type_list}).


Therefore, we have adopted ChatIE as the backbone of our automatic labeling module, with a slight modification that adds the trigger-related prompt template so that it can extract trigger words according to the event type. This allows for the extraction of trigger words based on the event type. For example, the following prompt template is used:
\texttt{When the event type of the given sentence above is ``<event-type>'', please recognize the corresponding trigger word. The trigger word is the word or phrase that most clearly expresses event occurrences.$\backslash n$Only answer the trigger word, no extra word. The trigger word is:}.

\section{Human Evaluation Procedure}
\label{sec:appendix human procedure}
All participants were gathered in a conference room and asked to sign a consent form before being introduced to the task and annotation criteria. They were then given model accounts to begin annotating without the use of external tools. Participants in the experimental group received ChatIE assistance. Once they finished annotating, they notified us using a communication tool, and we recorded the time spent on annotation. We evaluated metrics and conducted statistic analysis using a pre-written script.

\onecolumn
\section{Display}
\label{sec:appendix display}
\begin{figure*}[htp]
\centering
\includegraphics[width=1\linewidth]{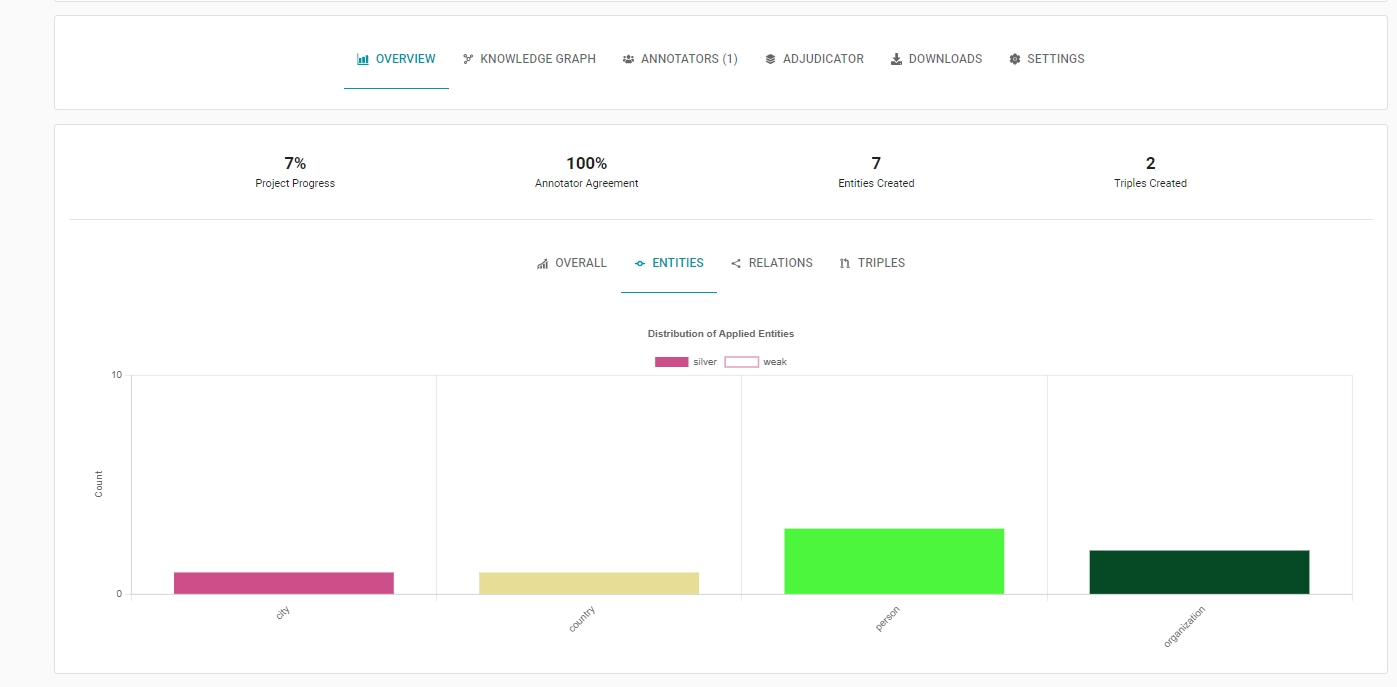}  
\caption{Overview of Dashboard.}
\label{fig:display-overview}
\vspace{-0.2em}
\end{figure*}

\begin{figure*}[htp]
\centering
\includegraphics[width=1\linewidth]{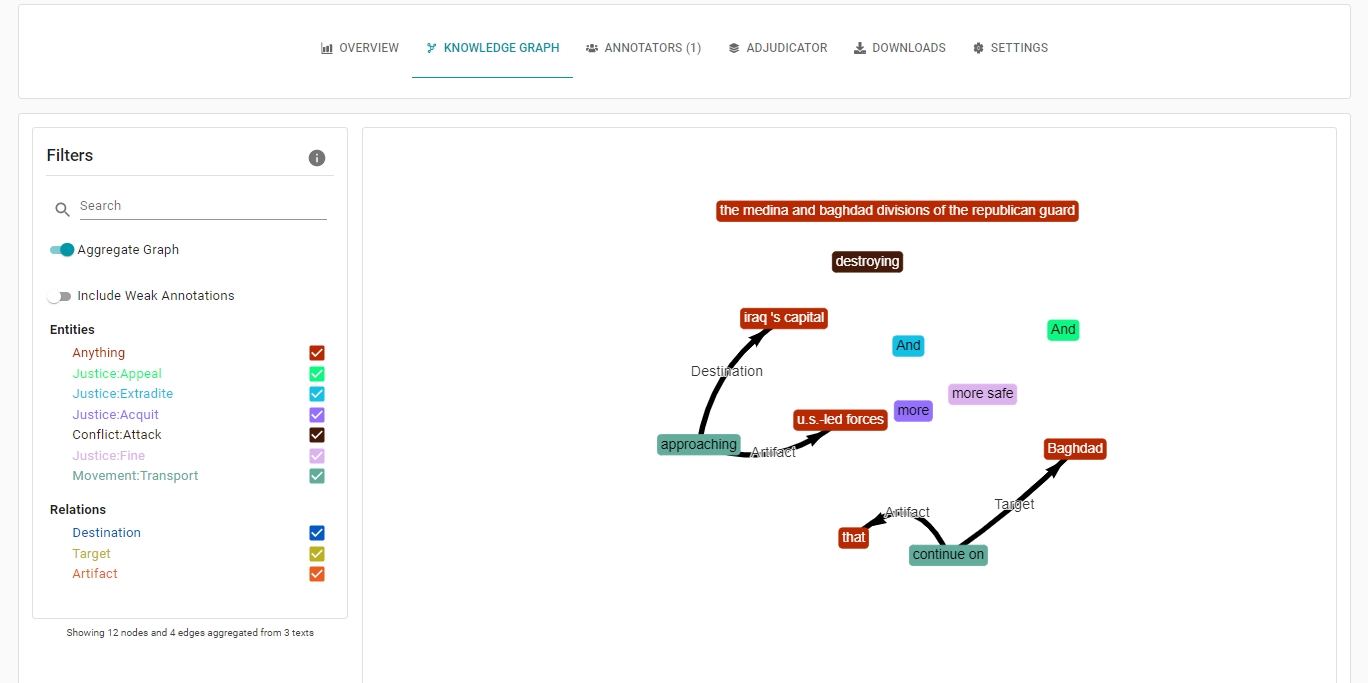}  
\caption{EKG display.}
\label{fig:display-ekg}
\vspace{-0.2em}
\end{figure*}

\begin{figure*}[htp]
\centering
\includegraphics[width=1\linewidth]{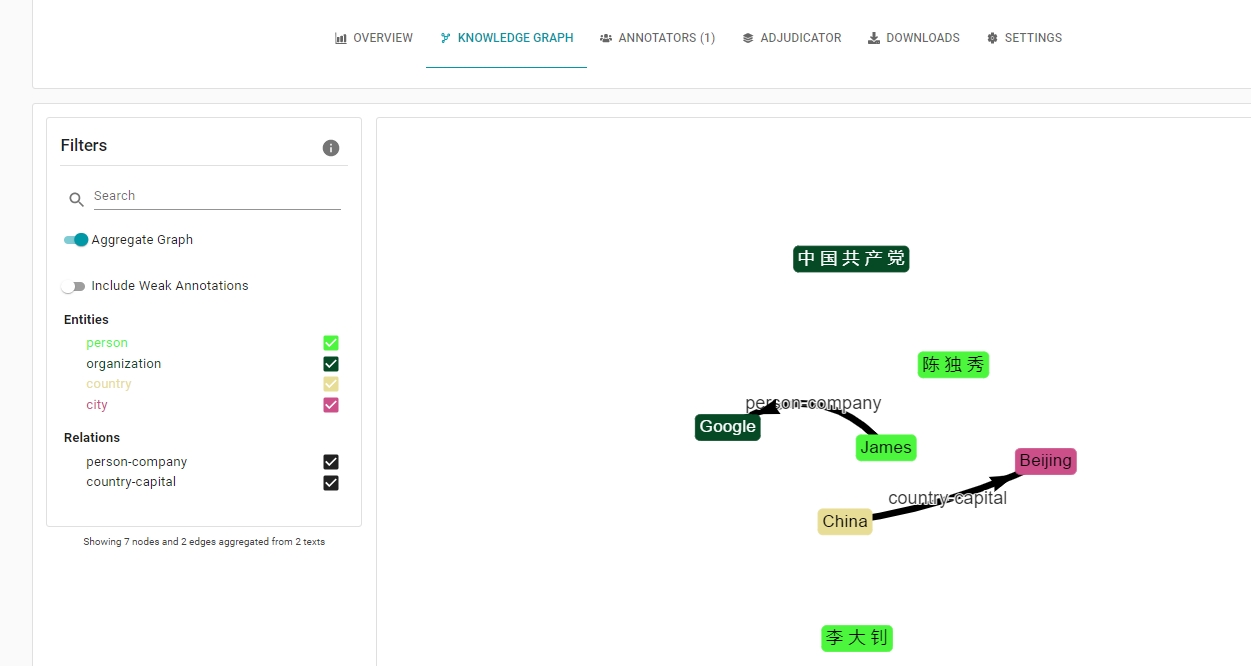}  
\caption{KG display.}
\label{fig:display-kg}
\vspace{-0.2em}
\end{figure*}

\begin{figure*}[htp]
\centering
\includegraphics[width=1\linewidth]{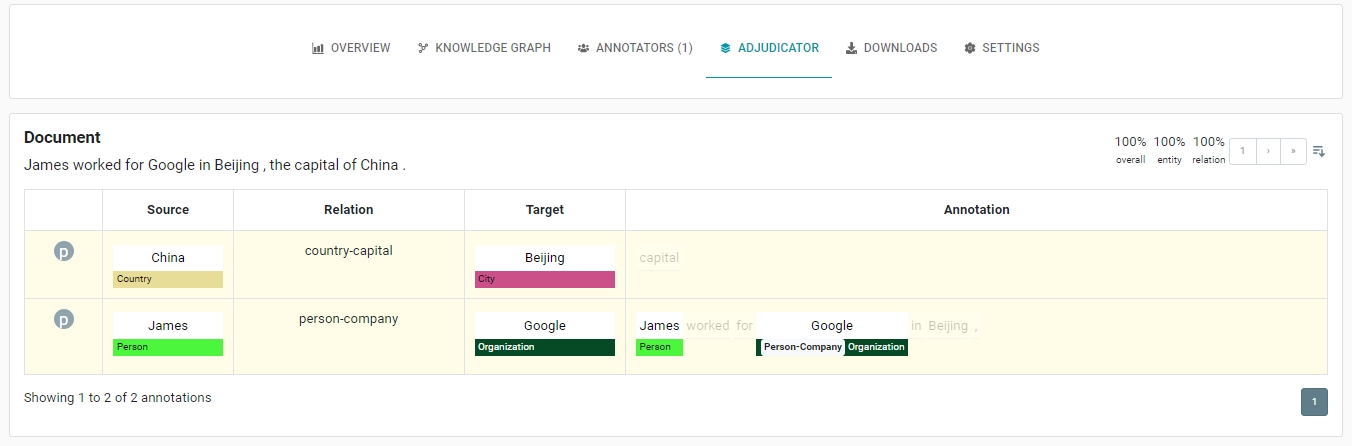}  
\caption{Double-checking display.}
\label{fig:display-adj}
\vspace{-0.2em}
\end{figure*}

\begin{figure*}[htp]
\centering
\includegraphics[width=1\linewidth]{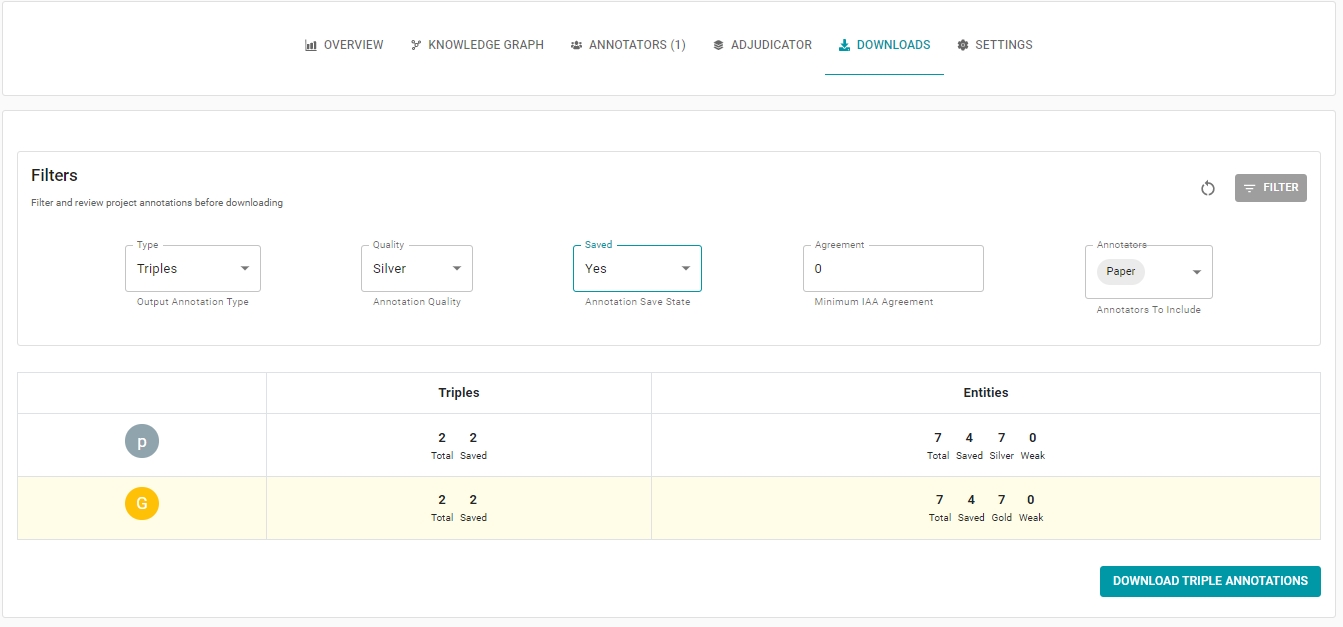}  
\caption{Download display.}
\label{fig:display-down}
\vspace{-0.2em}
\end{figure*}

Our toolkit offers a range of features to enhance the user-friendly display. Firstly, it displays annotation progress and counts the results across multiple dimensions, such as entity, relation, and triples (see Fig. \ref{fig:display-overview}). Secondly, users can view a KG or EKG and filter the results (see Fig. \ref{fig:display-ekg} and \ref{fig:display-kg}). Thirdly, the tool provides a double-checking function for each text (see Fig. \ref{fig:display-adj}). Finally, our tool supports the filtering and exporting functions (see Fig. \ref{fig:display-down}). Filter function includes saving, loading, quality filtering (accepted or suggested), and so on.

\section{Project Creation Process}
\label{sec:appendix p}

\begin{figure*}[htp]
\centering
\includegraphics[width=1\linewidth]{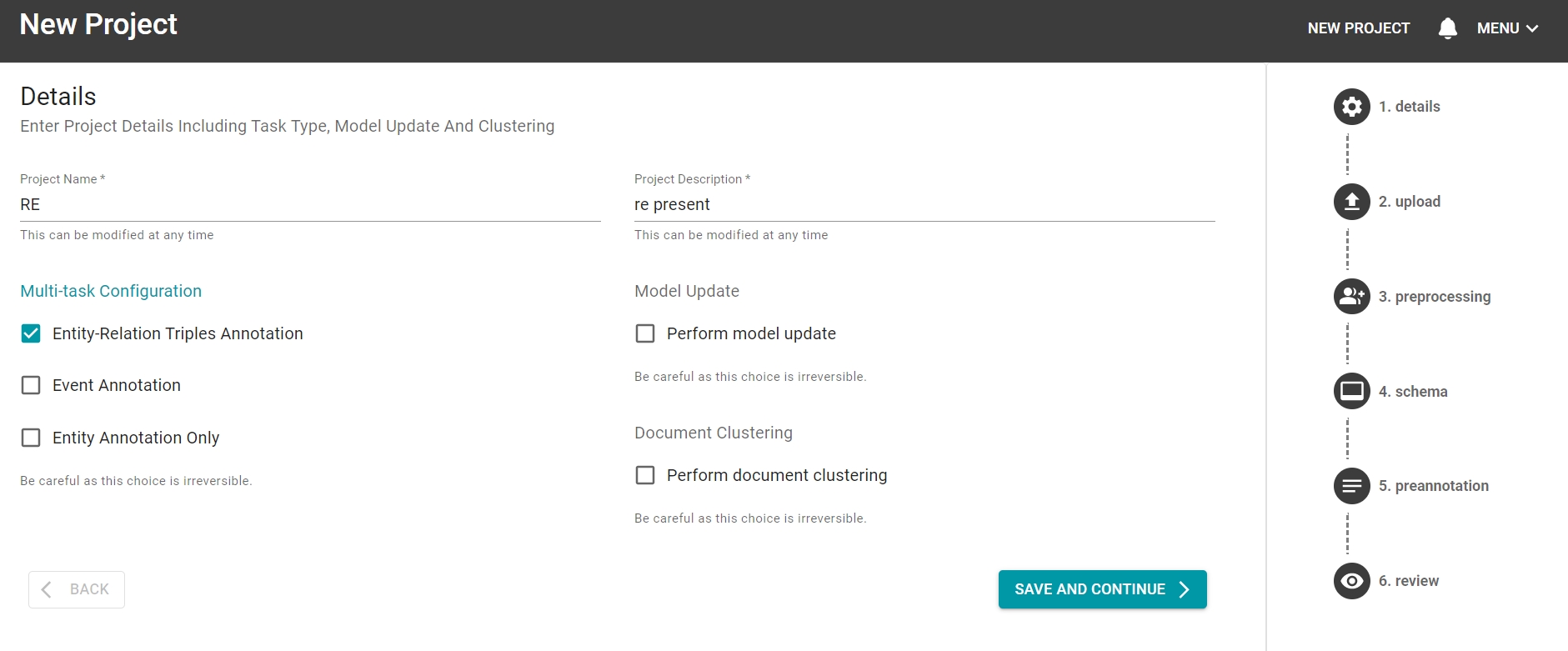}  
\caption{Detail of project creation.}
\label{fig:detail}
\vspace{-0.2em}
\end{figure*}

\begin{figure*}[htp]
\centering
\includegraphics[width=1\linewidth]{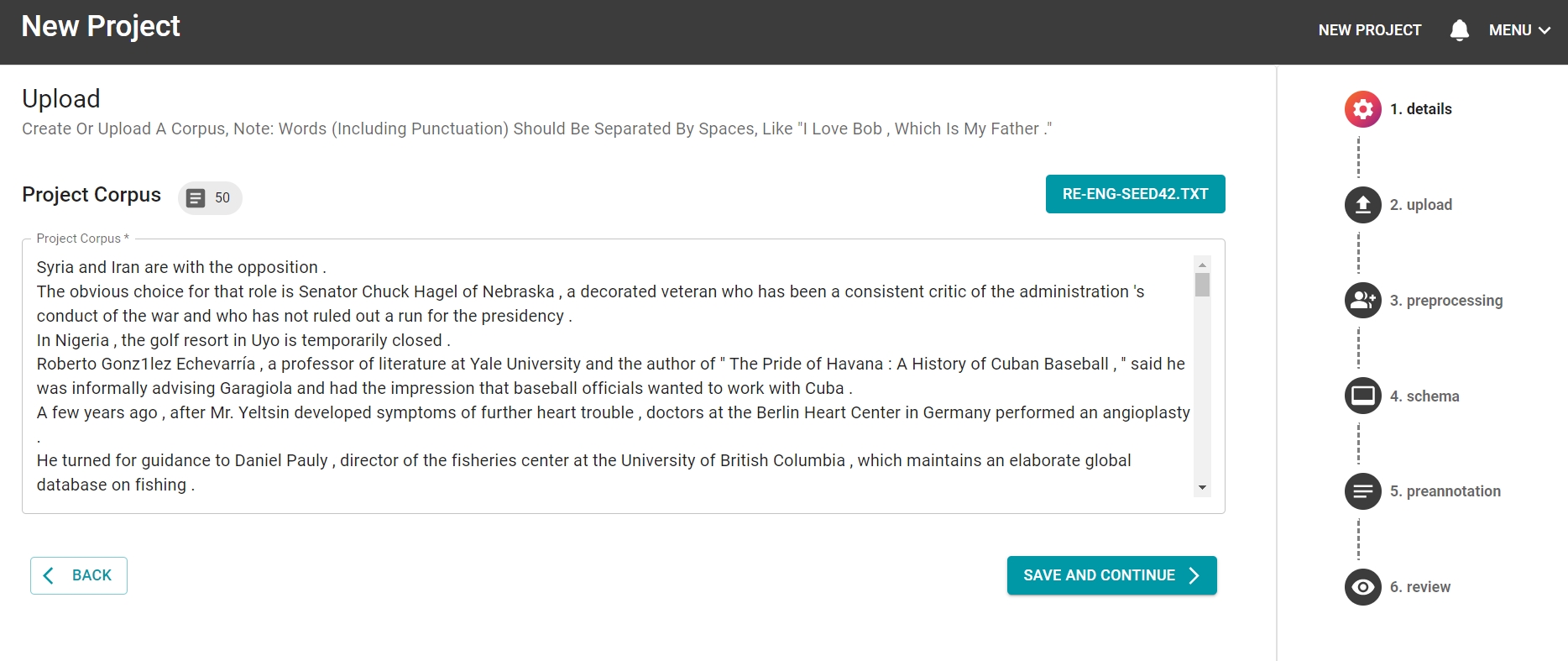}  
\caption{Uploading of project creation.}
\label{fig:upload}
\vspace{-0.2em}
\end{figure*}

\begin{figure*}[htp]
\centering
\includegraphics[width=1\linewidth]{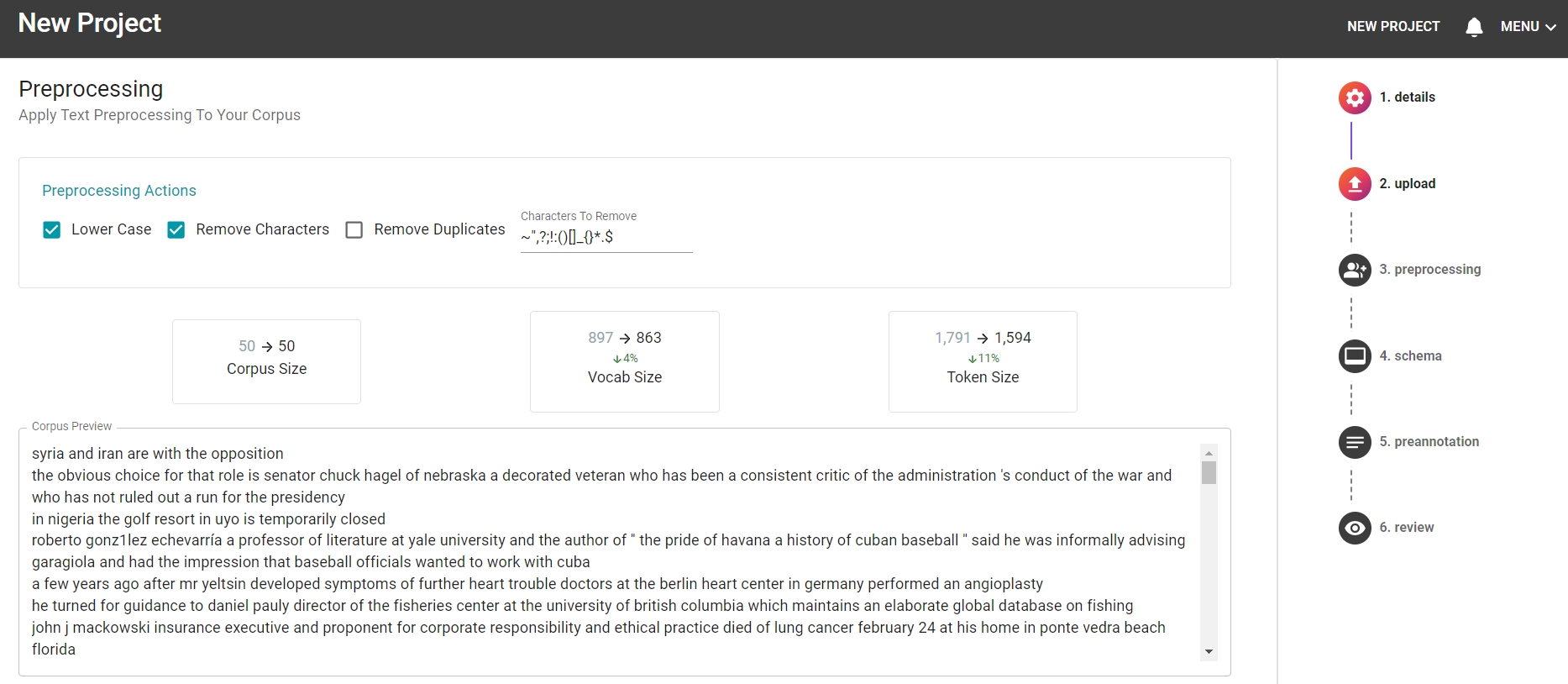}  
\caption{Preprocessing of project creation.}
\label{fig:preprocess}
\vspace{-0.2em}
\end{figure*}

\begin{figure*}[htp]
\centering
\includegraphics[width=1\linewidth]{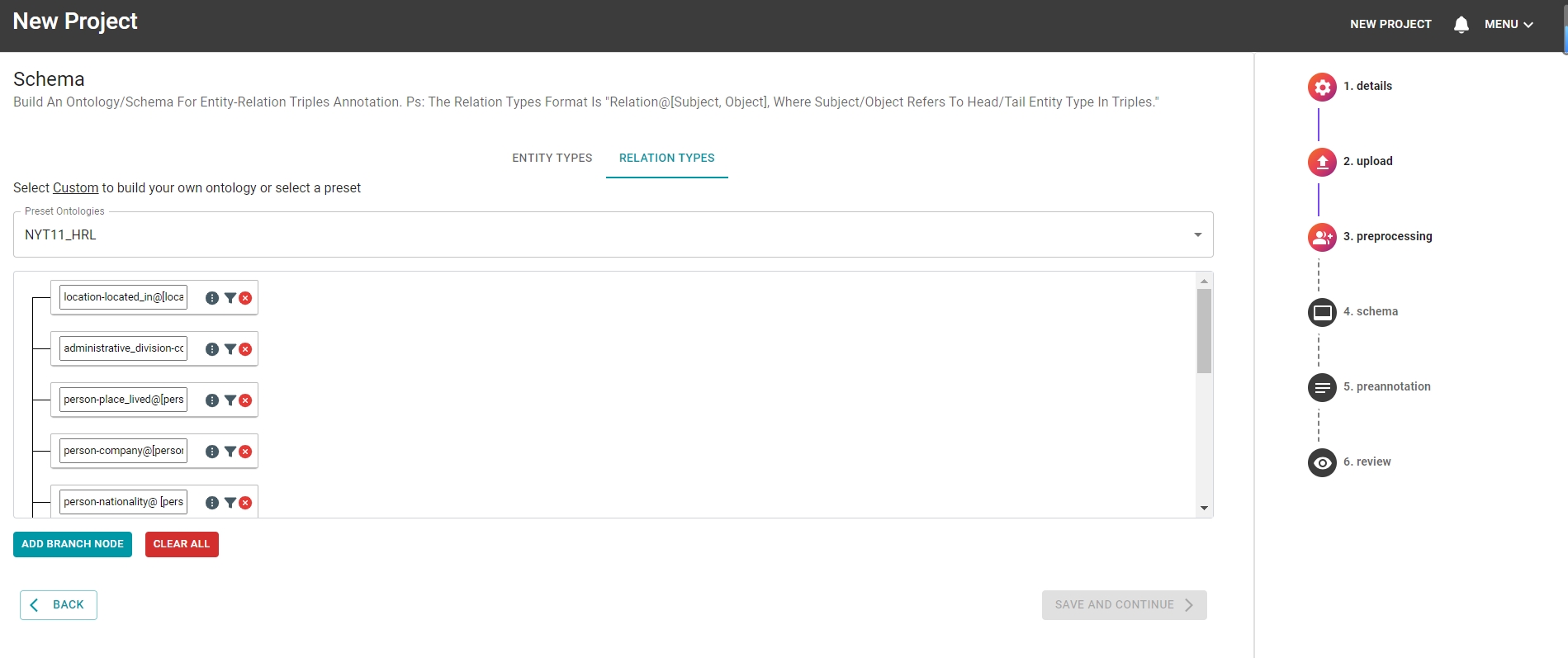}  
\caption{Scheme setup for RE of project creation.}
\label{fig:scheme for re}
\vspace{-0.2em}
\end{figure*}

\begin{figure*}[htp]
\centering
\includegraphics[width=1\linewidth]{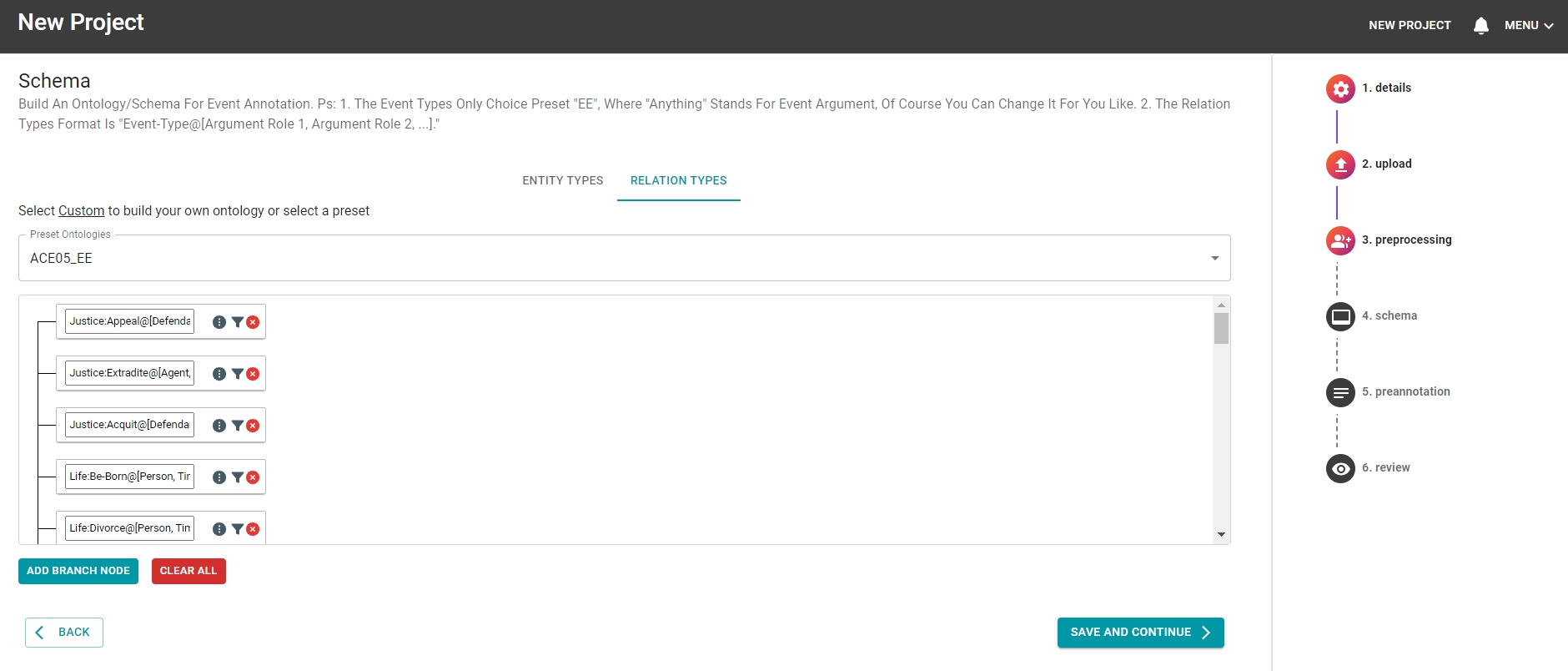}  
\caption{Scheme setup for EE of project creation.}
\label{fig:scheme for ee}
\vspace{-0.2em}
\end{figure*}

\begin{figure*}[htp]
\centering
\includegraphics[width=1\linewidth]{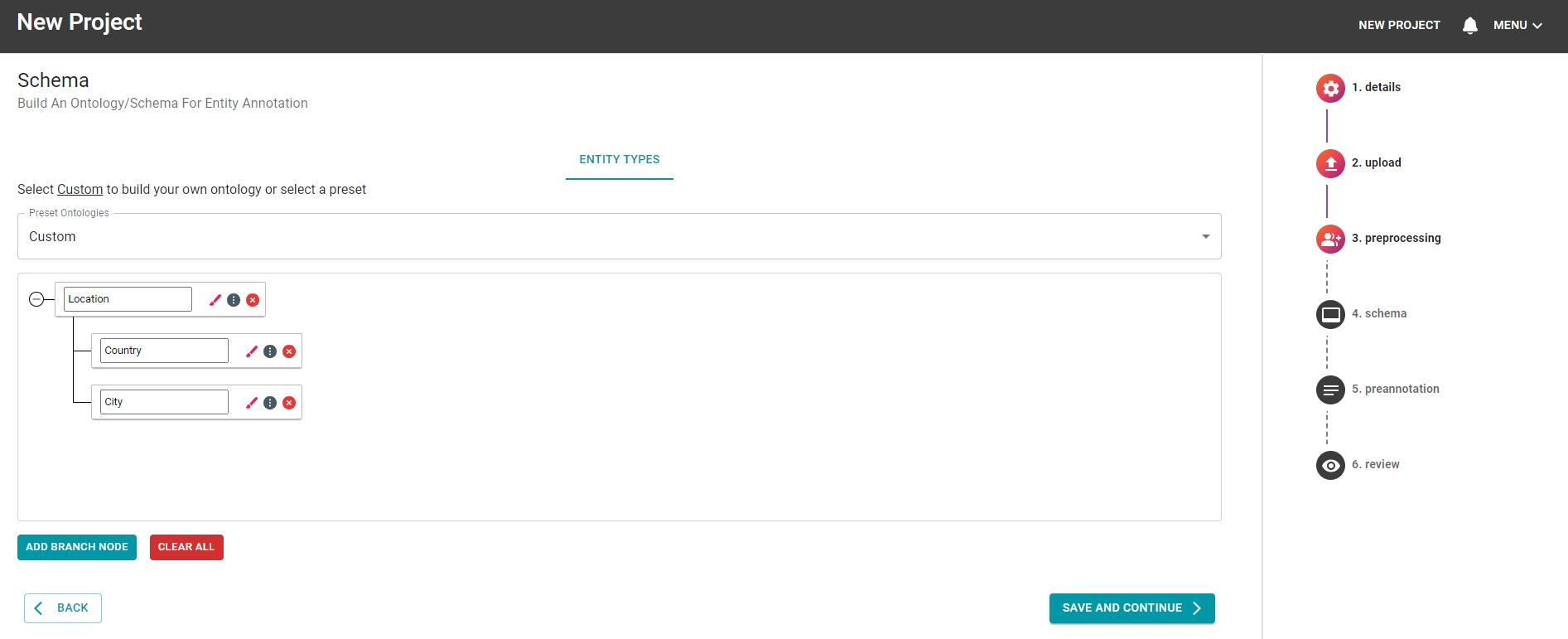}  
\caption{Hierarchical labels of project creation.}
\label{fig:hl}
\vspace{-0.2em}
\end{figure*}

\begin{figure*}[htp]
\centering
\includegraphics[width=1\linewidth]{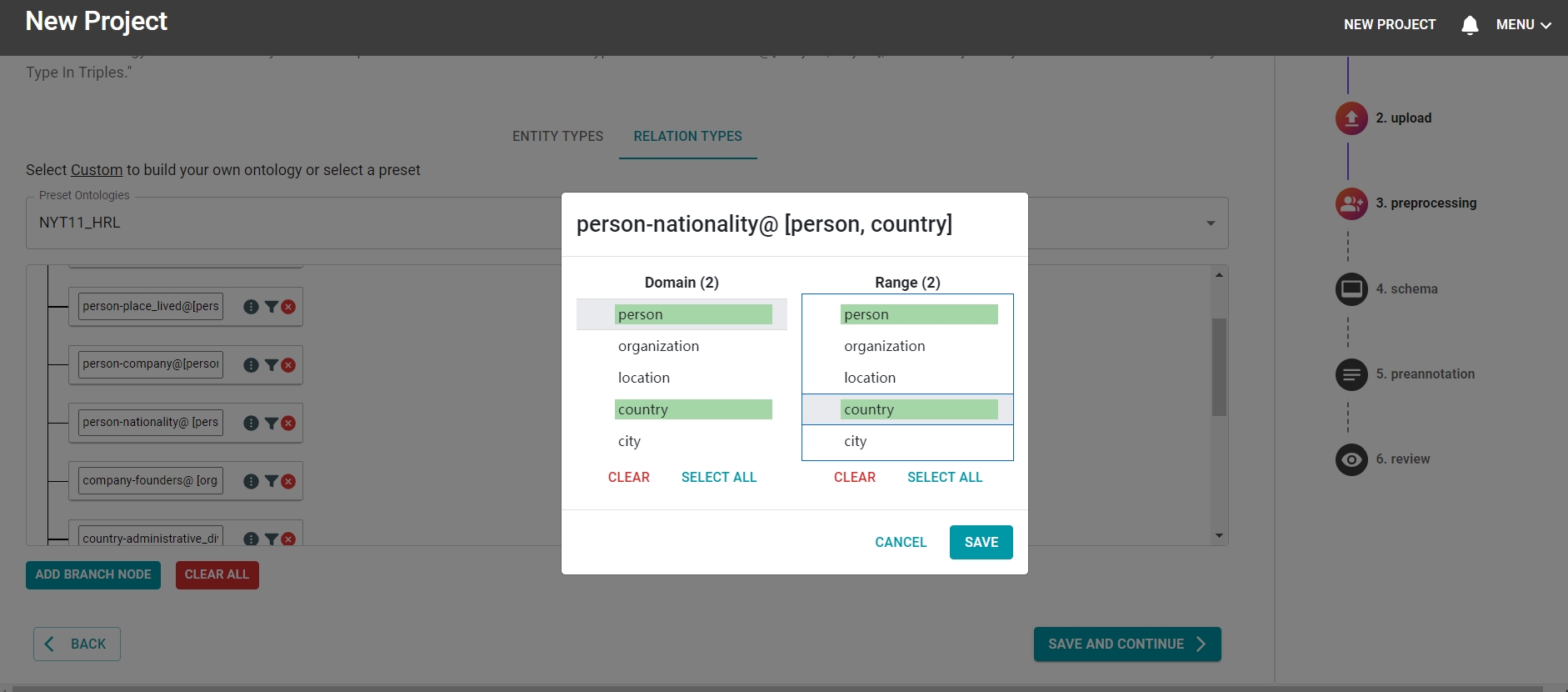}  
\caption{Relation constraint of project creation.}
\label{fig:rc}
\vspace{-0.2em}
\end{figure*}

\begin{figure*}[htp]
\centering
\includegraphics[width=1\linewidth]{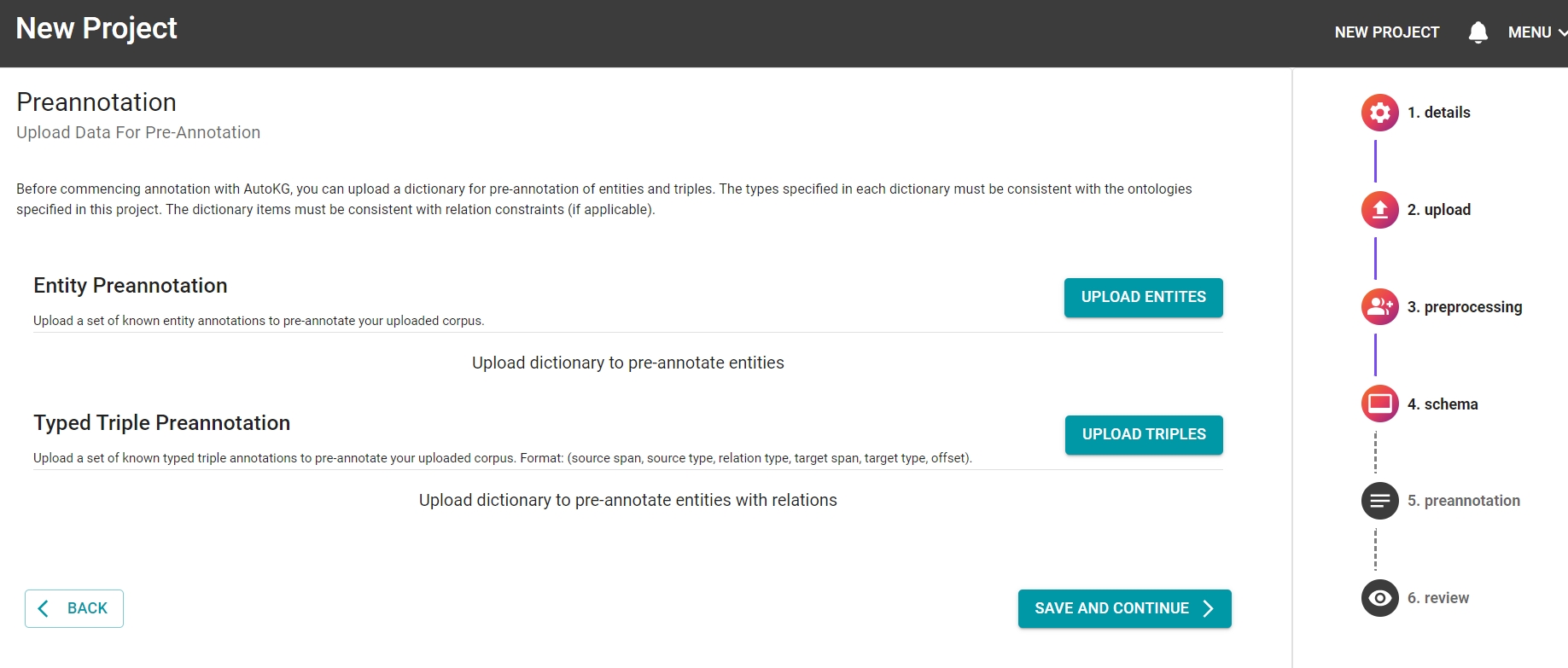}  
\caption{Preannotation of project creation.}
\label{fig:preannotation}
\vspace{-0.2em}
\end{figure*}

\begin{figure*}[htp]
\centering
\includegraphics[width=1\linewidth]{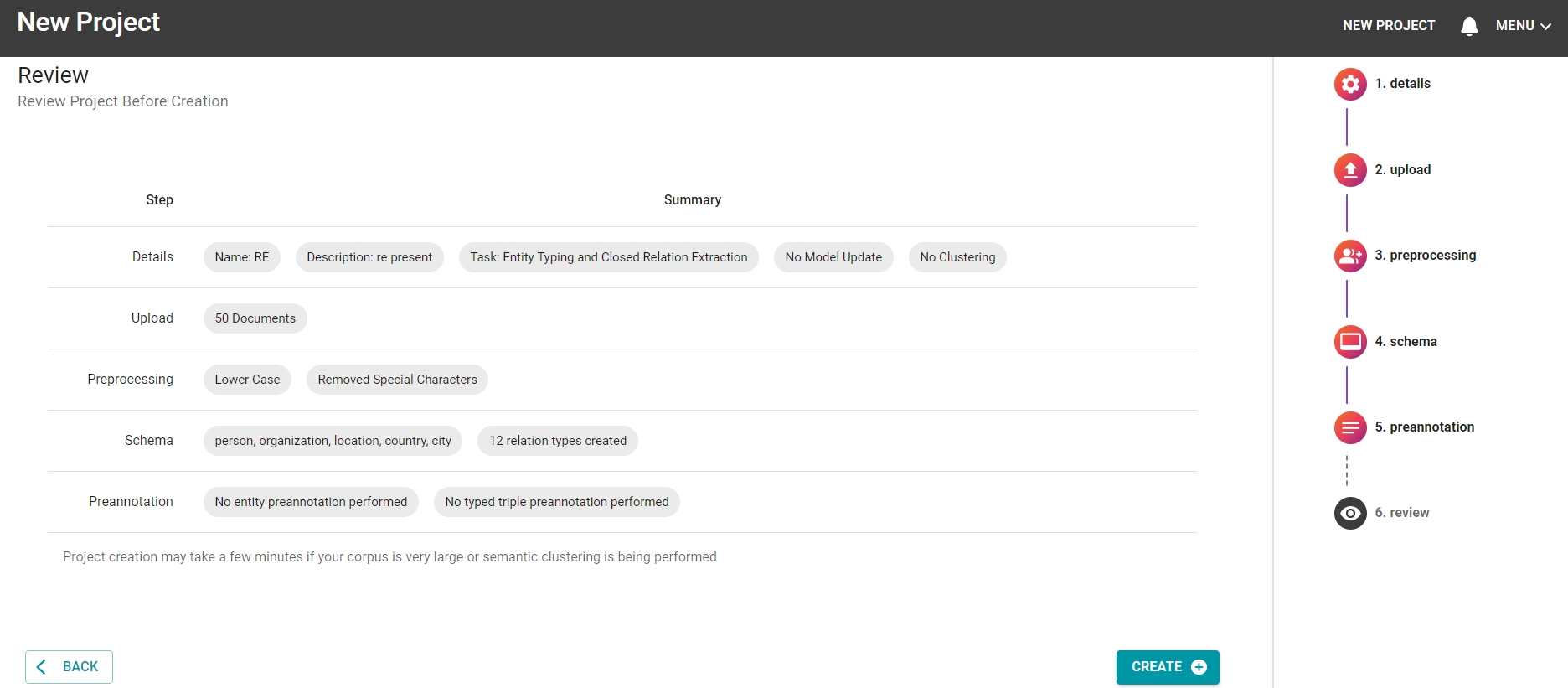}  
\caption{Review of project creation.}
\label{fig:review}
\vspace{-0.2em}
\end{figure*}

The project is created for each person on a dataset. The creation process can be divided into the following steps:

\begin{itemize}
    \item \emph{Configuration Setup: } Set up the details of the project: the name, the description, the configure of multi-task (NER, RE, EE), whether to perform \texttt{model update} (see Sec.\ref{sec:learnability}) and \texttt{text/document clustering} features (Fig.~\ref{fig:detail}).
    Clustering enables aggregation of texts with similar semantics (Fig. \ref{fig:mainui} A.IV) so that annotators can focus more on a certain class of concepts and thus increase productivity.
    Our toolkit implements cohesive clustering by encoding documents with SBERT~\citep{reimers-gurevych-2019-sentence} sentence embeddings.

    \item \emph{Uploading Data: } It supports keyboard input and uploading files (Fig.~\ref{fig:upload}).
    
    \item \emph{Pre-processing: } Pre-processing function includes character casing, specified-character removal, and text de-duplication (Fig.~\ref{fig:preprocess}).
    
    \item \emph{Scheme Setup: } Build an ontology/scheme for the current task. Users can choose from the preset ontology or \textbf{customize} their own scheme.
    For RE, the \texttt{RELATION TYPES} format is \texttt{relation@[subject, object]}, where subject/object refers to head/tail entity type in triples (Fig.~\ref{fig:scheme for re}).
    For EE, the \texttt{ENTITY TYPES} is filled with a pseudo token (namely \_), and the \texttt{RELATION TYPES} format is \texttt{event-type@[argument role 1, argument role 2, ...]} (Fig.~\ref{fig:scheme for ee}). It is worth noting that \model will complete the processing to convert \texttt{role} to \texttt{relation} and \texttt{event-type} to \texttt{entity-type} on the back-end.
    Unlike other IE annotation tools, our tool supports \textbf{\texttt{hierarchical labels}} (Fig.~\ref{fig:hl}) and \textbf{\texttt{relation constraint}} (Fig.~\ref{fig:rc}).
    The hierarchical labels facilitate the management of complex schemes.
    The relation constraint is a predetermination that a relationship can only occur between certain entity types. Consequently, this feature can narrow down the annotator's attention and improve the productivity and consistency of the annotators.

    \item \emph{Preannotation: } Users can choose to upload pre-annotated entities and relations of the current corpus. This can reduce annotation effort by pre-applying tags based on external resources such as gazetteers (Fig.~\ref{fig:preannotation}).
    
    \item \emph{Review: } Summarize the current project. Hence users can check and make changes (Fig.~\ref{fig:review}).
        
\end{itemize}

\begin{figure*}[htp]
\centering
\includegraphics[width=1\linewidth]{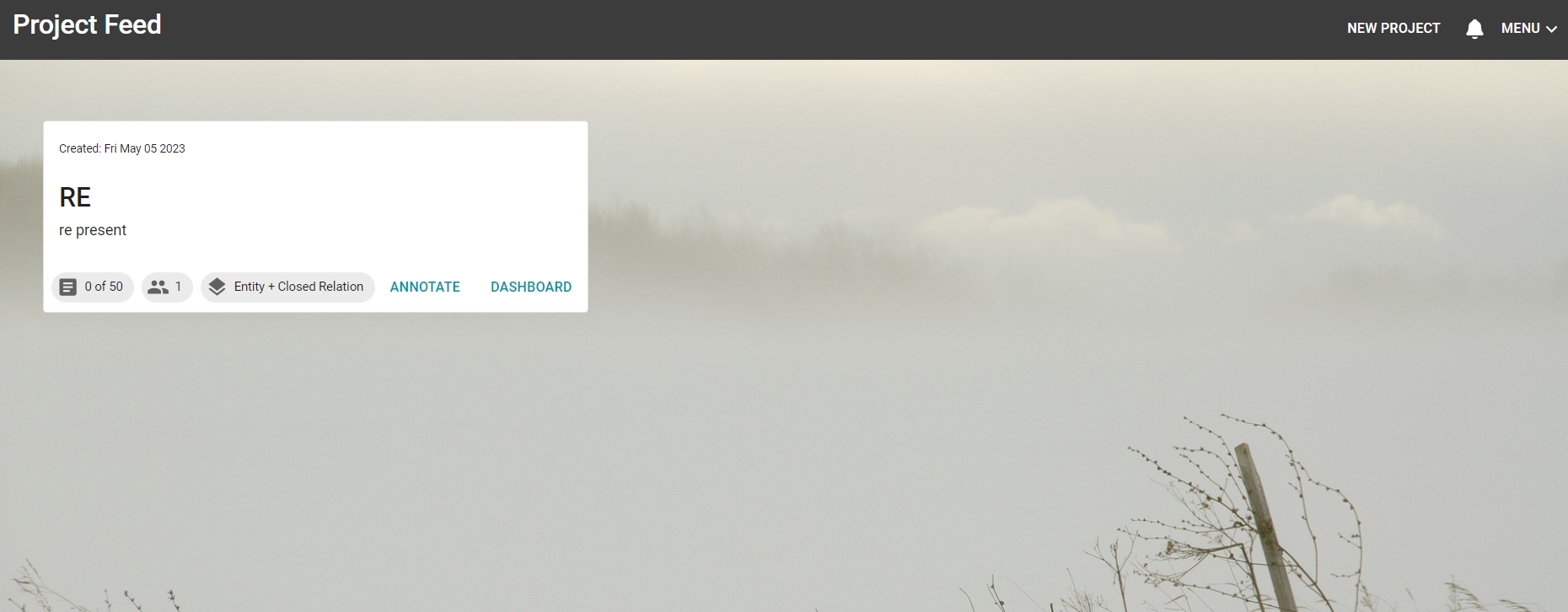}  
\caption{Feed of projects.}
\label{fig:feed}
\vspace{-0.2em}
\end{figure*}

Finally, when the user clicks the \texttt{CREATE} button, the project creation process is completed and will appear in the panel (Fig.~\ref{fig:feed}).

\end{document}